\documentclass[lettersize,journal]{IEEEtran}

\usepackage{amsmath,amssymb,mathtools,bm,empheq,amsthm}

\newtheorem{lemma}{Lemma}
\newtheorem{theorem}{Theorem}

\def\bal#1\eal{\begin{align}#1\end{align}} % align environment
\def\suml{\sum\limits}

\def\sg{\sigma}

\newcommand{\etal}{\textit{et al}., }
\newcommand{\ie}{\textit{i}.\textit{e}., }

\newcommand{\hbm}[1]{\hat{\bm{#1}}}

\newcommand{\br}[1]{\left[#1\right]} % left-right brackets
\newcommand{\pr}[1]{\left(#1\right)} % left-right parenthesis 
\newcommand{\cbr}[1]{\left\{#1\right\}} % left-right curly brackets
\DeclareMathOperator*{\argmin}{arg\,min} % argmin
\def\transp{\mathsf{T}} % tranpose symbol
\def\m{\mathbf}
\def\mc{\mathcal}
\def\R{\mathbb{R}}

\def\C{\mathbb{C}}

\def\ast{*}

\newcommand{\grad}[2]{\ensuremath{\nabla_{#2}#1}} % gradient operator
\newcommand{\norm}[2]{\ensuremath{\left\|#1\right\|_{#2}}}

\newcommand {\bbmtx}{\begin{bmatrix}} % begin matrix environment
\newcommand {\ebmtx}{\end{bmatrix}} % end matrix environment

\DeclareMathOperator*{\vctr}{vec} % vector
\newcommand{\vc}[1]{\vctr{\left(#1\right)}}

\newcommand{\deriv}[2]{\frac{\partial{#1}}{\partial{#2}}}

\DeclareMathOperator*{\diagonal}{diag} % diagonal
\newcommand{\diag}[1]{\diagonal\pr{#1}}
\DeclareMathOperator*{\trace}{tr} % diagonal
\newcommand{\tr}[1]{\trace\pr{#1}}

\DeclareMathOperator{\E}{\mathbb{E}}
\usepackage{graphicx}
\usepackage{amsmath,amssymb,amsfonts}
\usepackage{amsthm}
\usepackage{mathrsfs}
\usepackage{tabularx}
\usepackage{xcolor}
\usepackage{textcomp}
\usepackage{manyfoot}
\usepackage{booktabs}
\usepackage[ruled]{algorithm2e}
\SetKwRepeat{Do}{repeat}{until}
\usepackage{listings}
\usepackage{changepage}
\usepackage{caption}
\usepackage{multirow,tabularx}
\usepackage[export]{adjustbox}
\usepackage{calc}
\usepackage{stackengine}
\usepackage{hyperref}
\newtheorem{proposition}[theorem]{Proposition}
\usepackage{enumitem,kantlipsum}

\usepackage{amsmath,amsfonts}
\usepackage{algorithmic}
\usepackage{array}
\usepackage[caption=false,font=normalsize,labelfont=sf,textfont=sf]{subfig}
\usepackage{textcomp}
\usepackage{stfloats}
\usepackage{url}
\usepackage{verbatim}
\usepackage{graphicx}
\usepackage{cite}

\begin{document}

\title{Iterative Reweighted Least Squares Networks With Convergence Guarantees for Solving Inverse Imaging Problems}

\author{\begin{tabular}{@{}c@{}}
    Iaroslav Koshelev\\
    \href{mailto:koshelev.iaroslav@huawei.com}{koshelev.iaroslav@huawei.com}
  \end{tabular}%
  \quad 
  \begin{tabular}{@{}c@{}}
    and\\
     \quad
  \end{tabular}%
  \quad
  \begin{tabular}{@{}c@{}}
    Stamatios Lefkimmiatis\\
    \href{mailto:stamatios.lefkimmiatis@huawei.com}{stamatios.lefkimmiatis@huawei.com}
  \end{tabular}
  \thanks{I. Koshelev and S. Lefkimmiatis are with the Huawei Technologies Ltd, Noah’s Ark Lab.}\vspace{-0.5cm}}

\maketitle

\begin{abstract}
In this work we present a novel optimization strategy for image reconstruction tasks under analysis-based image regularization, which promotes sparse and/or low-rank solutions in some learned transform domain. We parameterize such regularizers using potential functions that correspond to weighted extensions of the $\ell_p^p$-vector and $\mc S_p^p$ Schatten-matrix quasi-norms with $0 < p \le 1$. Our proposed minimization strategy extends the Iteratively Reweighted Least Squares (IRLS) method, typically used for synthesis-based $\ell_p$ and $\mc S_p$ norm and analysis-based $\ell_1$ and nuclear norm regularization. We prove that under mild conditions our minimization algorithm converges linearly to a stationary point, and we provide an upper bound for its convergence rate. Further, to select the parameters of the regularizers that deliver the best results for the problem at hand, we propose to learn them from training data by formulating the supervised learning process as a stochastic bilevel optimization problem. We show that thanks to the convergence guarantees of our proposed minimization strategy, such optimization can be successfully performed with a memory-efficient implicit back-propagation scheme. We implement our learned IRLS variants as recurrent networks and assess their performance on the challenging image reconstruction tasks of non-blind deblurring, super-resolution and demosaicking. The comparisons against other existing learned reconstruction approaches demonstrate that our overall method is very competitive and in many cases outperforms existing unrolled networks, whose number of parameters is orders of magnitude higher than in our case. The code 
is available at \href{https://gitee.com/ys-koshelev/models/tree/lirls/research/cv/LIRLS}{this link}.
\end{abstract}

\begin{IEEEkeywords}
IRLS, $\ell_p$/$\mc{S}_p$ quasi-norms, Image Reconstruction, Recurrent Networks, Majorization-Minimization, Implicit Differentiation
\end{IEEEkeywords}

\section{Introduction}\label{sec:intro}
Inverse problems are ubiquitous and arise in a plethora of imaging applications ranging from digital photography and computer vision to biomedical imaging and remote sensing~\cite{Bertero1998, Szeliski2022}. In almost all the scenarios of practical interest, such problems are ill-posed~\cite{Tikhonov1963, Vogel2002}, meaning that the access to the measurements is not adequate by itself to guarantee a unique and stable solution. Nevertheless, by taking into account available prior knowledge related to certain characteristics of the solution space, it is still feasible to recover a meaningful approximation. In this case the recovered solution will be consistent to the observation model and at the same time will feature the desired properties, which are dictated by the available prior information. 

Among the existing strategies for dealing with inverse problems, variational methods are widely preferred in imaging applications and entail the use of modern regularization techniques~\cite{Benning2018}. Such image regularization methods are also directly linked to the maximum-a-posteriori (MAP) or penalized maximum likelihood estimators~\cite{Figueiredo2007}, which typically arise in the Bayesian framework. Under the variational treatment of inverse problems, a solution is obtained via the minimization of an objective function that consists of two main terms. The fist is the \textit{data-fidelity} term, which measures the proximity of the solution to the measurement and ensures that it is consistent with the forward (observation) model. The second is the 
\textit{regularization} term, whose role is to favor solutions that feature desired properties. The proper selection of this term is very important and plays a crucial role to the final result, as it determines both the quality of the reconstruction and the complexity of the recovery process.

A lot of effort has been put to the design of proper image regularizers for the past two decades. This has led to several regularization functionals, which are able to model different image properties that have been empirically observed and verified. In particular, it is well-documented that undistorted images typically have sparse responses when high-pass derivative-like filters are applied on them, such as the gradient~\cite{Rudin1992}, the Laplacian and the Hessian~\cite{Lefkimmiatis2012} and the discrete wavelet transform~\cite{Figueiredo2007}. The non-local self-similarity, which describes the repetition of localized patterns at distant locations across the image, and the correlation of the color image channels are other well-known properties of natural images that have been utilized in collaborative filtering approaches~\cite{Dabov2007, Buades2010, Lefkimmiatis2015bJ}. These last two properties lead to a linear dependence of image patches or features, which can be successfully enforced by employing a low-rank regularization on the responses of matrix-valued operators such as the Hessian, the structure tensor and the non-local structure tensor~\cite{Lefkimmiatis2013J,Lefkimmiatis2013Jb,Lefkimmiatis2015J, Lefkimmiatis2015bJ}.  Based on the above, both sparsity and low-rank promoting regularization methods have been studied in-depth and their use is supported both by solid mathematical theory~\cite{Donoho2006, Candes2008, Elad2010} as well as strong empirical results.  

There are two different types of regularization methods commonly used in inverse problems, namely the synthesis- and analysis-based ones. Synthesis-based regularization arises from the Basis Pursuit method proposed in the seminal paper in~\cite{Chen2001}. In this framework, the signal under recovery $\bm x \in \R^n$ is expressed as a linear combination of the so-called \textit{atoms}, which form an overcomplete dictionary $\bm D \in \R^{n \times m}, m \ge n$, such that $\bm x = \bm D \bm \gamma$. Then, a regularization approach is pursued w.r.t $\bm \gamma$, which is typically of much higher dimension than the primal signal of interest $\bm x$. On the other hand, analysis-based regularization utilizes a regularization operator that is directly applied to the signal of interest, essentially analysing its behaviour. Under this framework, a regularization operator $\bm G^{m \times n}$ transforms the signal $\bm x\in \R^{n}$ into an auxiliary features representation $\bm z = \bm G \bm x$, which in many cases is expected to be sparse or low-rank for a properly selected $\bm G$. For complete and under-complete transformations both synthesis- and analysis-based regularizers can be equivalent~\cite{Elad2007}. However, for the most interesting and common case where overcomplete representations are involved, the two types differ and for image reconstruction tasks analysis-based regularization usually leads to better reconstruction performance~\cite{Selesnick2009}. 

Apart from the regularization operator $\bm G$, another important component in analysis-based sparse and low-rank promoting priors is the potential function that enforces the transformed features to exhibit the desired property. In this case, the $\ell_0$ norm and the matrix rank are ideal candidates to serve such role. Unfortunately, the use of either of these penalties leads to NP-hard optimization problems, which are intractable when dealing with high-dimensional signals such as images and videos. To overcome this issue, their closest convex relaxations are typically considered. As a result the $\ell_1$ norm is selected to approximate the $\ell_0$ pseudo-norm ~\cite{Rudin1992,Figueiredo2007}, while the nuclear norm is used to approximate the matrix rank ~\cite{Lefkimmiatis2013J, Lefkimmiatis2015J}. More recently, it has been advocated that non-convex penalties such as the $\ell_p^p$ vector and $\mc S_p^p$ Schatten-matrix quasi-norms better enforce the sparsity and the low-rank property, respectively, and lead to improved image reconstruction results~\cite{Chartrand2007, Lai2013, Candes2008c, Gu2014, Liu2014, Xie2016, Kummerle2021}. Extensions of such penalties involving their weighted versions have also been proposed, which have enriched the representation ability of both sparse and low-rank priors in image denoising and background subtraction tasks~\cite{Gu2014, Xie2016}.   

Due to their non-convex and non-smooth nature, the minimization of objective functions involving such regularizers is very challenging. Basic techniques like gradient descent-based methods are known to deliver unsatisfactory results due to their slow convergence speed and instabilities caused by the improper choice of hyperparameters (e.g. step size). Other minimization strategies that have been proposed include variable-splitting-based (VS) methods such as the Half-Quadratic splitting~\cite{Nikolova2005} and the Alternating Direction Method of Multipliers (ADMM)~\cite{Boyd2011} as well as proximal-gradient methods like the Fast Iterative Shrinkage Algorithm (FISTA)~\cite{Beck2009b}, or even a combination of both~\cite{Zuo2013}. For the first category, the tuning of algorithmic parameters is required and their proper selection is crucial for achieving a satisfactory convergence rate. On the other hand, proximal-gradient methods require the computation of the proximal operator for the regularizer~\cite{Boyd2004} at each iteration, which cannot be analytically derived in the majority of the cases. The Iteratively Reweighted Least Squares (IRLS) is another minimization strategy that has been successfully used in dealing with inverse problems when $\ell_p^p$~\cite{Daubechies2010} and $\mc S_p^p$~\cite{Mohan2012} regularizers~\cite{Lai2013} are employed. IRLS does not involve any parameters that need to be tuned and demonstrates fast convergence, with the original optimization problem being transformed into a specific sequence of easy-to-solve quadratic sub-problems. These are important benefits that favour its usage in various large-scale imaging tasks, where either sparse~\cite{Zhou2014} or low-rank~\cite{Fornasier2011} priors are employed. 

In the past, in order to employ such low-rank and sparsity promoting priors in image reconstruction tasks, the specific parametric form of the regularizer had to be manually selected. While several of the existing hand-crafted regularizers, with the most prominent example being the Total Variation (TV)~\cite{Rudin1992}, have been used in various inverse problems with relative success, the modern paradigm of deep-learning has clearly demonstrated that learned regularizers can lead to far superior performance than classical regularization techniques. In particular, feed-forward convolution neural networks (CNNs) trained to perform denoising~\cite{Lefkimmiatis2017, Zhang2017b, Romano2017, Lefkimmiatis2018} can implicitly act as an image prior. Based on this observation, several methods employing denoising CNNs were later proposed to deal with general image reconstruction tasks, where the degradation operator is other than identity, including demosaicking~\cite{Kokkinos2018, Kokkinos2019}, burst denoising/demosaicking~\cite{Kokkinos2019C}, deblurring~\cite{Dong2021} and super-resolution~\cite{Zhang2020}. However, one potential drawback of such networks arises from their strength: due to the huge number of involved trainable parameters, these networks have a large capacity and tend to overfit to the training data. As a result, their performance can degrade significantly when applied to data unseen during training. Another drawback of such methods is that they do not explicitly model any of the well-studied image properties, as those mentioned earlier, hence their performance is hard to analyze and interpret~\cite{Monga2021}. 

Utilizing learning procedures for classical image reconstruction algorithms is very challenging, as they typically involve an optimization process and hence are of a recurrent nature. The unrolling strategy is the most popular way to learn such recurrent networks, where the parameters are updated using the Backpropagation Through Time (BPTT) algorithm. Given that the memory constraints usually do not allow running the forward recurrent iterations until convergence, their cutoff is performed based on a small predefined value. Since the restoration quality is highly correlated with the iterations number~\cite{Kokkinos2019}, other approaches such as truncated BPTT~\cite{Robinson1987} were proposed to increase the depth of the networks for the purpose of improving their inference results. Recently, another class of effectively infinite depth networks, namely Deep Equilibrium Models (DEQ) has been introduced~\cite{Bai2019} and applied for image reconstruction~\cite{Gilton2021}. Such networks employ the implicit function theorem for efficient backpropagation without the need to unroll. However, the main drawback of general DEQ models is that they lack convergence guarantees for their forward iterative process, which is a prerequisite for training using implicit backpropagation. It is also worth noting that long before the introduction of DEQs, similar strategies have been proposed for discriminative learning of Markov Random Fields~\cite{Kegan2009} and later for learning sparsity enforcing analysis and synthesis priors under the framework of bilevel optimization~\cite{Peyre2011}, a field that continues to develop up to the present day~\cite{Ghosh2022}.

In our previous work~\cite{Lefkimmiatis2023}, we proposed a novel IRLS-type algorithm for the minimization of two families of analysis-based regularizers that involve as their potential functions smooth and weighted extensions of the vector $\ell_p^p$ and the matrix Schatten $\mc S_p^p$ quasi-norms with $p \in (0; 1]$. This allowed us to efficiently deal with various general image reconstruction problems under sparse and/or low-rank constraints. Furthermore, we have coupled our method with a proposed stochastic training strategy based on the implicit function theorem, and achieved very competitive results surpassing other state-of-the-art reconstruction approaches. Here, we extend our previous work in the following ways: \textbf{(1)} We provide a formal proof for the linear convergence of our algorithm to a finite stationary point and derive its convergence rate. \textbf{(2)} For the color image case, we consider a sparsity-enforcing prior which leads to a novel reconstruction network that surpasses both our previously reported results and the current state-of-the-art approaches.
\textbf{(3)} We discuss in detail practical issues that arise during the the implementation and training of our networks and describe how we efficiently address them. 
\textbf{(4)} We provide an extensive line of experimental results that has allowed us to draw a direct performance comparison between the sparsity and low-rank promoting priors for the same reconstruction problems and benchmarks.

\section{Problem Formulation and Image Recovery}
Hereafter, considering a general image reconstruction problem, we will use the following notation. With $\bm y\!\in\!\R^{m\cdot c'}$ we denote the observation vector that is related to the multichannel underlying image $\bm x\!\in\!\R^{n\cdot c}$ of $c$ channels, that we seek to recover. Their relation is dictated by the degradation process (forward model), which is described by a linear operator $\bm A: \R^{n\cdot c} \rightarrow \R^{m\cdot c'}$ corresponding to the impulse response of the imaging device, and a noise vector $\bm n\!\in\!\R^{m\cdot c'}$ modeling all the approximation errors of the forward model as well as the measurement noise:
\bal
\bm y = \bm A\bm x +\bm n.
\label{eq:fwd_linear_model}
\eal
In this work we assume that $\bm n$ consists of i.i.d samples drawn from a Gaussian distribution of zero mean and variance $\sigma_{\bm n}^2$. Despite of the simplicity of the adopted observation model, it is widely used in the literature and can accurately describe a plethora of practical problems. Indeed, by properly selecting the operator $\bm A$ in Eq.~\eqref{eq:fwd_linear_model}, one can describe various inverse problems such as denoising, deblurring, demosaicking, inpainting, super-resolution, MRI reconstruction, etc. Under this forward model, an estimate of the latent image $\bm x^\ast$ can be recovered as the minimizer of the primal objective function:
\bal
\mc J\pr{\bm x} =\tfrac{1}{2\sigma_{\bm n}^2}\norm{\bm y-\bm A\bm x}{2}^2 + \mc R\pr{\bm x}.
\label{eq:objective}
\eal
where $\mc R\!:\!\R^{n\cdot c}\!\rightarrow\!\R_+\!=\!\cbr{x\in\R | x \ge 0}$ denotes the employed regularizer (image prior). 

\subsection{Sparse and Low-Rank Analysis-Based Image Priors}
Most of the existing analysis-based image regularizers presented in the literature can be summarized with the following generic formulation:
\bal
\mc R\pr{\bm x} = \sum\limits_{i=1}^\ell \phi\pr{\bm G_i\bm x},
\eal
where $\bm G\!:\!\R^{n\cdot c}\!\rightarrow\!\R^{\ell\cdot d}$, $\bm G_i\!=\!\bm M_i\bm G$, $\bm M_i\!=\!\bm I_{d} \otimes \bm e_i^\transp$ with $\otimes$ denoting the Kronecker product, and $\bm e_i$ is the unit vector of the standard $\R^\ell$ basis. Further, $\phi\!:\!\R^{d}\!\rightarrow\!\R_+$ is a potential function that penalizes the response of the $d$-dimensional transform-domain feature vector,  $\bm z_i\!=\!\bm G_i\bm x\!\in\!\R^d$. For the reasons discussed in the introduction, in this work we focus on sparsity and low-rank promoting regularizers and we consider two expressive parametric forms for the potential function $\phi\pr{\cdot}$, which correspond to weighted and smooth extensions of the $\ell_p^p$ and the Schatten matrix $\mc S_p^p$ quasi-norms  with $0<p\le 1$, respectively. The first one is a sparsity-promoting penalty, defined as:
\bal
\phi_{sp}\pr{\bm z;\bm w, p} = \suml_{j=1}^d\bm w_j\pr{\bm z_j^2+\gamma}^{\frac{p}{2}},\, \bm z, \bm w\in\R^d, 
%\mbox{ with } 0 < p \le 1 \mbox{ and } \gamma > 0,
\label{eq:sparse_prior}
\eal
while the second one is a low-rank (spectral-domain sparsity) promoting penalty, defined as:
\bal
&\phi_{lr}\pr{\bm Z;\bm w, p} = \suml_{j=1}^r\bm w_j\pr{\bm \sg_j^2\pr{\bm Z}+\gamma}^{\frac{p}{2}}, \bm Z\in\R^{m\times n}, \bm w\in\R_+^r, \nonumber\\ &\quad\quad\mbox{ with } r=\min\pr{m, n}.
\label{eq:low_rank_prior}
\eal
In both definitions $\gamma$ is a small fixed constant that ensures the smoothness of the penalty functions. Moreover, in Eq.~\eqref{eq:low_rank_prior} $\bm\sg\pr{\bm Z}$ denotes the vector with the singular values of $\bm Z$ sorted in decreasing order, while the weights $\bm w$ are restricted to be sorted in increasing order. The intuition behind the opposite ordering between the singular values and the weights is that in order to better promote low-rank solutions, the smaller singular values of the matrix have to be penalized more than its larger ones, since these are the ones that ideally need to attain the zero value. Next, we define our proposed sparse and low-rank promoting image priors as:
\begin{subequations}
   \begin{tabularx}{\hsize}{@{}XX@{}}
     \begin{equation}
	\mc R_{sp}\pr{\bm x} = \suml_{i=1}^{\ell}\phi_{sp}\pr{\bm z_i;\bm w_i, p}\label{eq:reg_sp},
     \end{equation}&
     \begin{equation}
       \mc R_{lr}\pr{\bm x} = \suml_{i=1}^{\ell}\phi_{lr}\pr{\bm Z_i;\bm w_i, p}\label{eq:reg_lr},
     \end{equation}
   \end{tabularx}
\label{eq:regularizers}
 \end{subequations}
where in Eq.~\eqref{eq:reg_sp} $\bm z_i\!=\!\bm G_i\bm x\!\in\!\R^d$, while in Eq.~\eqref{eq:reg_lr} $\bm Z_i\!\in\!\R^{c\times q}$ is a matrix whose dependence on $\bm x$ is expressed as: $\vc{\bm Z_i}\!=\!\bm G_i\bm x\!\in\!\R^d$, with $d\!=\!c\!\cdot\! q$. The $j$-th row of $\bm Z_i$ is formed by the $q$-dimensional feature vector $\bm Z_i^{\pr{j,:}}$ extracted from the image channel $\bm x^{j}, j=1,\!\ldots,\! c$.

\subsection{Image Recovery via Majorization-Minimization}\label{sec:MM}
As in our prior work~\cite{Lefkimmiatis2023}, we adopt the same majorization-minimization (MM)~\cite{Hunter2004} framework to minimize the overall objective function in Eq.~\eqref{eq:objective}, which may include any of the considered sparsity or low-rank promoting regularizers. Specifically, instead of minimizing the objective function $\mc J\pr{\bm x}$ directly, we follow an iterative procedure where each iteration consists of two steps: (\textbf{a}) selection of a surrogate function that serves as a tight upper-bound of the original objective (\emph{majorization-step}) and (\textbf{b}) computation of a current estimate of the solution by minimizing the surrogate function (\emph{minimization-step}).
Based on the above, the iterative algorithm for solving the minimization problem $\bm x^\ast=\argmin_{\bm x}\mc J\pr{\bm x}$ takes the form:
%\bal
$\bm x^{k+1}=\argmin_{\bm x}\mc Q\pr{\bm x;\bm x^k}$,
%\eal
where $\mc Q\pr{\bm x;\bm x^k}$ is the majorizer of the objective function $\mc J\pr{\bm x}$ at some point $\bm x^k$, satisfying the two conditions:
\bal
\label{eq:MM_properties}
\mc Q\pr{\bm x;\bm x^k} \ge \mc J\pr{\bm x}, \forall \bm x, \bm x^k\,\,\mbox{and}\,\,
\mc Q\pr{\bm x^k;\bm x^k} = \mc J\pr{\bm x^k}.
\eal
Given these two properties of the majorizer, it can be easily shown that the iterative minimization of $\mc Q\pr{\bm x;\bm x^k}$ leads to the monotonic decrease of the objective function $\mc J\pr{\bm x}$~\cite{Hunter2004}. In fact, to ensure this, it only suffices to find a $\bm x^{k+1}$ that decreases the value of the majorizer, \ie $\mc Q\pr{\bm x^{k+1};\bm x^k} \le \mc Q\pr{\bm x^k;\bm x^k}$. Moreover, given that both $\mc Q\pr{\bm x;
\bm x^k}$ and $\mc J\pr{\bm x}$ are bounded from below, we can state that the sequence $\cbr{\mc J\pr{\bm x^k}}$ converges as $k \to \infty$. This result is powerful by itself, but more assumptions should be made to ensure that the sequence $\cbr{\bm x^k}$ converges to the stationary point $\bm x^*$ of $\mc J\pr{\bm x}$. Specifically, if the following two additional conditions are satisfied 
\bal
\label{eq:additional_MM_properties}
\mc Q\pr{\bm x;\bm x^k} \ \mbox{is continuous wrt} \ \bm x, \bm x^k \quad \mbox{and} \nonumber \\ \nabla_{\bm x} \mc Q\pr{\bm x;\bm x^k}|_{\bm x^k = \bm x} = \nabla_{\bm x} \mc J\pr{\bm x},
\eal
then every limit point of the sequence $\cbr{\bm x^k}$ is a stationary point of the original problem~\cite{Razaviyayn2013}. However, even this result does not provide a complete picture about the convergence behaviour, such as the rate of convergence, as well as a guarantee that $\cbr{\bm x^k}$ does not approach infinity. In this work we extend our previous results and provide formal proofs of the convergence behaviour and the rate of convergence, which are presented later.

A favorable type for the majorizer of our image prior $\mc R\pr{\bm x}$ would be a quadratic one. This stems from the fact that the data fidelity term of the objective function $\mc J\pr{\m x}$ in Eq.~\eqref{eq:objective} is itself quadratic and, thus, the overall majorizer would also be of the same form. Such a majorizer is highly desirable since it is amenable to efficient minimization by solving the corresponding normal equations. Below we provide two lemmas that allow us to design such tight quadratic upper bounds both for the sparsity-promoting~\eqref{eq:reg_sp} and the low-rank promoting~\eqref{eq:reg_lr} regularizers. We note, that these results are provided in a different but still equivalent form than the one used in our previous work in~\cite{Lefkimmiatis2023}. The reason for this, is that the current chosen form better facilitates the derivation of the convergence proof of our algorithm. 
\begin{lemma}\label{lem:lp_upper_bound}
Let $\bm x,\, \bm y\in\R^n$ and $\bm w\in\R_+^n$.  The weighted-$\ell_p^p$ function $\phi_{sp}\pr{\bm x;\bm w, p}$ defined in Eq.~\eqref{eq:sparse_prior} can be upper-bounded as:
\bal
\phi_{sp}\pr{\bm x;\bm w, 
p} \le& \phi_{sp}\pr{\bm y;\bm w, p} + \frac{p}{2} \bm x^\transp\bm W_{\bm y}\bm x \nonumber\\&- \frac{p}{2} \bm y^\transp \bm W_{\bm y} \bm y,\, \forall\,\bm x,\bm y%\in\R^n
\label{eq:lp_majorizer}
\eal
where $\bm W_{\bm y} = \diag{\bm w}\br{\bm I\circ \pr{\bm y\bm y^\transp+\gamma\bm I}}^{\frac{p-2}{2}}$ and $\circ$ denotes the Hadamard product. The equality in \eqref{eq:lp_majorizer} is attained when $\bm x=\bm y$.
\end{lemma}

\begin{lemma}\label{lem:sp_upper_bound}
Let $\bm X, \bm Y\!\in\!\R^{m\times n}$ and $\bm \sg\!\pr{\bm Y}, \bm w\!\in\!\R_+^r$ with $r\!=\!\min\pr{m, n}$. The vector $\bm \sg\!\pr{\bm Y}$ holds the singular values of $\bm Y$ in decreasing order while the elements of $\bm w$ are sorted in increasing order. The weighted-Schatten-matrix function $\phi_{lr}\pr{\bm X;\bm w, p}$ defined in Eq.~\eqref{eq:low_rank_prior} can be upper-bounded as:
\bal
\phi_{lr}\pr{\bm X;\bm w, p}\le& \phi_{lr}\pr{\bm Y;\bm w, p} + \frac{p}{2} \trace \pr{\bm X^\transp \bm W_{\bm Y} \bm X} \nonumber \\&-\frac{p}{2} \tr{\bm Y^\transp \bm W_{\bm Y} \bm Y},\, \forall\,\bm X,\bm Y
\label{eq:Sp_majorizer}
\eal
where $\bm W_{\bm Y} =  \bm U\diag{\bm w}\bm U^\transp\pr{\bm Y\bm Y^\transp+\gamma\bm I}^{\frac{p-2}{2}}$ and $\bm Y=\bm U\diag{\bm\sg\pr{\bm Y}}\bm V^\transp$, with $\bm U\in\R^{m\times r}$ and $\bm V\in\R^{n\times r}$. The equality in \eqref{eq:Sp_majorizer} is attained when $\bm X=\bm Y$.
\end{lemma}

Now, having access to the above tight upper-bounds we can readily obtain the quadratic majorizers for both of our regularizers, $\mc Q_{reg}$, as:

\begin{subequations}\vspace{-0.05\hsize}
   \begin{tabularx}{\hsize}{@{}X@{}}
	\bal
	  \mc Q_{sp}\pr{\bm x; \bm x^k} = \tfrac{p}{2}\suml_{i=1}^\ell\bm z_i^\transp\bm W_{\bm z_i^k}\bm z_i,
	\eal\\\vspace{-1.2cm}
	\bal
	  \mc Q_{lr}\pr{\bm x; \bm x^k}= \tfrac{p}{2}\suml_{i=1}^\ell\trace\pr{\bm Z_i^\transp\bm W_{\bm Z_i^k}\bm Z_i},
	\eal
	\end{tabularx}
\label{eq:majorizers}
\end{subequations}
where $\bm z_i$ and $\bm Z_i$ are defined as in Eq.~\eqref{eq:regularizers}, $\bm W_{\bm z_i^k}, \bm W_{\bm Z_i^k}$ are defined in Lemmas~\ref{lem:lp_upper_bound} and \ref{lem:sp_upper_bound}, respectively, while we have ignored all the constant terms that do not depend on $\bm x$. In both cases, by adding the majorizer of the regularizer, $\mc Q_{reg}\pr{\bm x;\bm x^k}$, to the quadratic data fidelity term, we obtain the overall majorizer $\mc Q\pr{\bm x;\bm x^k}$, of the objective function $\mc J\pr{\bm x}$. It is also possible to move one step further and consider an augmented overall majorizer of the form:
\bal
\tilde{\mc Q}\pr{\bm x;\bm x^k} = \mc Q\pr{\bm x;\bm x^k} + \frac{\delta}{2}\norm{\bm x-\bm x^k}{2}^2,
\eal
which also satisfies all the requirements of the MM framework, since it still holds that:
\bal
\tilde{\mc Q}\pr{\bm x;\bm x} = \mc J\pr{\bm x}\quad\mbox{and}\quad \tilde{\mc Q}\pr{\bm x^k;\bm x} \ge \mc J\pr{\bm x}\,\,\forall \bm x,\bm x^k.
\eal
As it will become clear next, one reason for utilizing $\tilde{\mc Q}$ instead of $\mc Q$ is that it ensures the uniqueness of the solution in each iteration. Another purpose that it serves is that it forces the current estimate not to differ significantly from the one of the previous iteration. Both of these properties play an important role to the stability of the training stage of the recurrent networks that we discuss in Section~\ref{sec:LIRLS_implementation}.

Now, since our majorizer is quadratic, we obtain the $(k\!\!+\!\!1)$-th update of the MM iteration by solving the normal equations:
\bal
&\bm x^{k+1} = \pr{\bm S^k + \alpha\bm I}^{-1}\bm b^k = \label{eq:normal_eq}\\ &\pr{\bm A^\transp\bm A + p\!\cdot\!\sg_{\bm n}^2\suml_{i=1}^\ell\bm G_i^\transp\bm W_i^k\bm G_i + \alpha\bm I}^{-1}\pr{\bm A^\transp\bm y +\alpha\bm x^k},\nonumber
\eal
where $\alpha\!=\delta\sg^2_{\bm n}$, $\bm W_i^k\!=\!\bm W_{\bm z_i^k}\!\in\!\R^{d\times d}$ for $\mc R_{sp}\pr{\bm x},$ and 
$\bm W_i^k\!=\!\bm I_q\!\otimes\!\bm W_{\bm Z_i^k}\!\in\!\R^{c\cdot q\times c\cdot q}$ for $\mc R_{lr}\pr{\bm x}$. We note that due to the presence of the term $\alpha\bm I$, the system matrix $\bm S^k + \alpha \bm I$ is non-singular and thus invertible, which justifies our choice of using the augmented majorizer $\tilde{\mc Q}$ instead of $\mc Q$.
Based on all the above, the minimization of $\mc J\pr{\bm x}$, incorporating any of the two regularizers of Eq.~\eqref{eq:regularizers}, boils down to solving a sequence of re-weighted least squares problems, where the weights $\bm W_i^k$ of the current iteration are updated according to the solution of the previous one. Given that our regularizers in Eq.~\eqref{eq:regularizers} include the weights $\bm w_i\ne\bm 1$, our proposed algorithm generalizes the IRLS methods introduced by~\cite{Daubechies2010} and~\cite{Mohan2012}, which only consider the case where $\bm w_i = \bm 1$ and have been successfully applied in the past on sparse and low-rank recovery problems, respectively.

\begin{figure}
    \centering
    \includegraphics[width=\linewidth]{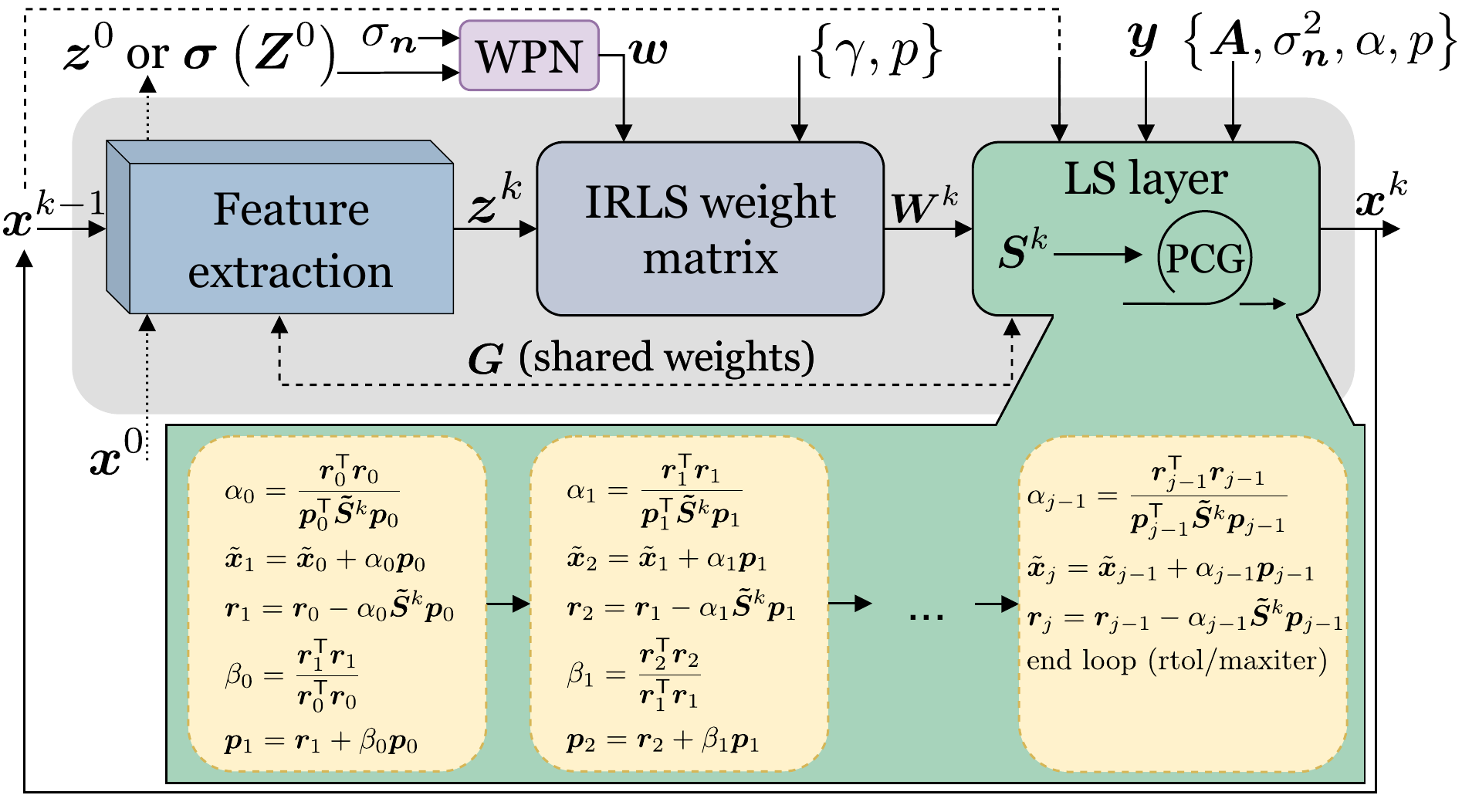}
    \caption{The proposed LIRLS recurrent architecture.}
    \label{fig:LIRLS_net}
\end{figure}

While Eqs.~\eqref{eq:MM_properties} are not sufficient by themselves to guarantee the convergence of the proposed iterative algorithm to a finite fixed point or a stationary point of $\mc J\pr{\bm x}$, in our previous work we have empirically observed that this is still the case. Below we provide a new theoretical result, which concludes that the algorithm indeed converges linearly to the finite stationary point for both sparse and low-rank priors. We formalize this in Prop.~\ref{prop:convergence}, and provide its proof in Appendix \ref{sec:prop_proof}. We note that this new result provides the complete analysis of the convergence behaviour of our algorithm.
\begin{proposition}\label{prop:convergence}
The iterative algorithm given by Eq.~\eqref{eq:normal_eq}: 
\begin{enumerate}[wide, labelwidth=!, labelindent=0pt]
    \item Converges to a finite fixed point $\bm x^*$, which is also a stationary point of $\mc J\pr{\bm x}$.
    \item Sufficiently close to the local minima the convergence is linear with a rate of at least 
    \bal
    &\mu_{\mc J}^{ub} = 1 - \nonumber \\ &\frac{\sigma_{\bm n}^2 \lambda_{\min}\pr{ \bm H_{\mc J\pr{\bm x^*}}}}{\norm{\bm A}{2}^2 + p \sigma_{\bm n}^2 \suml_{i=1}^\ell \norm{\bm G_i}{2}^2 \begin{cases} \max_j{\bm w_{ij} \pr{{\bm z_i^*}_j^2 + \gamma}^{\frac{p-2}{2}}} \\ \max_j{\bm w_{ij} \pr{\bm \sigma_j^2\pr{\bm Z_i^*} + \gamma}^{\frac{p-2}{2}}}\end{cases}\!\!\!\!\!\! + \alpha}, \label{eq:proposition}
    \eal
    where $\bm H_{\mc J\pr{\bm x^*}}$ denotes the Hessian matrix of $\mc J\pr{\bm x}$ at point $\bm x^*$ and the upper part of the cases in the denominator of Eq.~\eqref{eq:proposition} corresponds to the sparsity enforcing priors, while the bottom one to the low-rank enforcing priors, respectively. 
\end{enumerate}
\end{proposition}

\section{Learning Explicit Priors via Recurrent Reconstruction Networks}
To deploy the proposed IRLS algorithm, we have to specify both the regularization operator $\bm G$ and the parameters $\bm w=\cbr{\bm w_i}_{i=1}^{\ell}, p$ of the potential functions in Eqs.~\eqref{eq:sparse_prior} and~\eqref{eq:low_rank_prior}, which constitute the image regularizers of Eq.~\eqref{eq:regularizers}. Manually selecting their values, with the goal of achieving satisfactory reconstructions, can be a cumbersome task. Thus, instead of explicitly defining them, we pursue the idea of finding their optimal values from training data. Under this strategy, we can formulate the following bilevel optimization problem:
\bal
\bm \theta^\ast = \argmin_{\bm \theta} \E_{\pr{\bm x^\textrm{gt}, \bm y} \in \mc D} \mc L\pr{\bm x^{\textrm{gt}}, \bm x^\ast\pr{\bm \theta, \bm y}}, \nonumber \\ \textrm{s.t.} \ \bm x^\ast\pr{\bm \theta} = \argmin_{\bm x} \mc J\pr{\bm x, \bm \theta}
\label{eq:bilevel}
\eal
with  $\bm \theta\!=\!\cbr{\bm G,\bm w, p}$ denoting the parameters of the regularizers and $\mc L$ being the loss function. The upper level is a classical problem of supervised learning, where the loss function $\mc L$ is minimized across the data distribution $\mc D = \cbr{\pr{\bm x^\textrm{gt}, \bm y}}$ of ground truth and degraded pairs to match the true solution $\bm x^{\textrm{gt}}$ and the output $\bm x^\ast$ from the lower optimization problem. The latter is hence an inference problem, and contrary to Eq.~\eqref{eq:objective}, here we specifically indicate the dependence of the loss function on the parameters we intend to learn. In the described bilevel optimization formulation, the parameters $\bm \theta$ tie together both the upper and the lower level subproblems in such a way that they can not be transformed into a single optimization problem.

As described in the previous section, the solution of the lower optimization can be efficiently computed from our proposed majorization-minimization approach, and it would be convenient to represent it as an iterative procedure of the form $\bm x^{k+1}\!=\!f_{\bm\theta}\pr{\bm x^k; \bm y}$. Considering our goal of learning the parameters $\bm \theta\!=\!\cbr{\bm G,\bm w, p}$, such an iterative algorithm can be viewed as a recurrent neural network (RNN) $f_{\bm\theta}\pr{\bm x^k; \bm y}$. This observation gives rise to our learned IRLS-based reconstruction networks (LIRLS). Their three main components are: (\textbf{a}) A feature extraction layer that accepts as input the current reconstruction estimate, $\bm x^k$, and outputs the feature maps $\cbr{\bm z_i^k = \bm G_i\bm x^k}_{i=1}^\ell$. (\textbf{b}) The weight module that acts on $\bm z_i^k$ and the parameters $\bm w_i$ to construct the weight matrix $\bm W_i^k$ based on the results of Lemmas~\ref{lem:lp_upper_bound},\ref{lem:sp_upper_bound}, which is part of the system matrix $\bm S^k$ in Eq.~\eqref{eq:normal_eq}. (\textbf{c}) The least-squares (LS) layer whose role is to produce a refined reconstruction estimate, $\bm x^{k+1}$, as the solution of Eq.~\eqref{eq:normal_eq}. The overall architecture of LIRLS is shown in Fig.~\ref{fig:LIRLS_net}.

\vspace{-.03\hsize}\subsection{Stochastic Training With Implicit Backpropagation}
In image reconstruction problems the large size of the paired dataset $\mc D$ from Eq.~\eqref{eq:bilevel} coupled with the high dimensionality of $\bm x$ does not practically allow the direct optimization of the expectation $\E_{\pr{\bm x^\textrm{gt}, \bm y} \in \mc D}$. Thankfully, the resemblance of the lower optimization problem from Eq.~\eqref{eq:bilevel} to a recurrent neural network suggests that it is possible to overcome this issue by conducting the upper level optimization similarly to stochastic training of neural networks using backpropagation. 

We note that the fixed point $\bm x^\ast$ that is achieved upon convergence of IRLS is in fact the implicit function given by the first order optimality conditions $\nabla_{\bm x^\ast} \mc J\pr{\bm x^\ast, \bm \theta} = \bm 0$. Considering the form of our objective function from Eq.~\eqref{eq:objective} and the form of IRLS iterations, as shown in Eq.~\eqref{eq:normal_eq}, it is straightforward to translate this condition to:
\bal
g\pr{\bm x^\ast, \bm \theta}\equiv \bm x^\ast - f_{\bm\theta}\pr{\bm x^\ast;\bm y}=\bm S^\ast\pr{\bm x^\ast,\bm\theta}\bm x^\ast-\bm A^\transp\bm y = \bm 0, 
\label{eq:convergence_cond}
\eal
where we explicitly indicate the dependency of $\bm S^\ast$ on $\bm x^\ast$ and $\bm\theta$. To update the network's parameters during training,  we have to compute the gradients $\grad{\mc L\pr{\bm x^\ast}}{\bm\theta}=\grad{\bm x^\ast}{\bm\theta}\grad{\mc L\pr{\bm x^\ast}}{\bm x^\ast}$.
Now, if we differentiate both sides of Eq.~\eqref{eq:convergence_cond} w.r.t $\bm\theta$, then we get:
\bal
\deriv{g\pr{\bm x^\ast, \bm\theta}}{\bm\theta} + \deriv{g\pr{\bm x^\ast, \bm\theta}}{\bm x^\ast}\deriv{\bm x^\ast}{\bm\theta} =\bm 0 \Rightarrow \nonumber \\ \grad{\bm x^\ast}{\bm\theta} = - \grad{g\pr{\bm x^\ast, \bm\theta}}{\bm\theta}\pr{\grad{g\pr{\bm x^\ast, \bm\theta}}{\bm x^\ast}}^{-1}.
\eal

Thus, we can now compute the gradient of the loss function w.r.t the network's parameters as:
\bal
\grad{\mc L\pr{\bm x^\ast}}{\bm\theta}=-\grad{g\pr{\bm x^\ast, \bm\theta}}{\bm\theta}\bm v, 
\label{eq:backward_update}
\eal
where $\bm v$ is obtained as the solution of the linear problem $\grad{g\pr{\bm x^\ast, \bm\theta}}{\bm x^\ast}\bm v\!\!=\!\!\grad{\mc L\pr{\bm x^\ast}}{\bm x^\ast}$ and all the necessary auxiliary gradients can be computed via automatic differentiation. The above result allows us to learn the sparsity and low-rank promoting regularizers without having to restrict the overall effective depth of LIRLS or save any intermediate results that would significantly increase the memory requirements. The detailed procedure that is used to perform both the forward and backward passes is presented in Algorithm~\ref{alg:IBP}. 
\begin{algorithm}[h]
\footnotesize
 \SetAlgoCaptionSeparator{\unskip:}
 \SetAlgoSkip{}
 \SetAlgoNoEnd 
 \SetAlgoInsideSkip{}
 \DecMargin{1cm}
\SetKwInput{NI}{Inputs}
\SetKwInput{Np}{Input parameters}
\SetKwBlock{Fwd}{Forward Pass}{}
\SetKwBlock{Bwd}{Backward Pass}{}
\NI{$\bm x_0$: initial solution, $\bm y$: degraded image, $\bm A$: degradation operator}
\Np{$\bm\theta=\left\{ \bm G, \bm w, p\right\}$: network parameters, $\sigma_{\bm n}^2$, $\alpha$, $\gamma$}
\Fwd{
Initialize: $k = 0$\;
\Do{the convergence criterion is satisfied}{
\begin{enumerate}[wide, labelwidth=!, labelindent=0pt]
\item Compute the feature maps $\cbr{\bm z_i^k = \bm G_i\bm x^k}_{i=1}^\ell$  $\pr{\bm Z^k_i\!=\!\vc{\bm z^k_i} \textrm{for the low-rank case}}$.\\

\item Compute the updated $\bm W_i^k$ weight matrices based on the\\current estimate  $\bm x^k$:
\begin{itemize}
\footnotesize
    \item Sparse case: $\bm W_i^k = \diag{\bm w_i}\br{\bm I\circ \pr{\bm z^k_i {\bm z^k_i}^\transp+\gamma\bm I}}^{\frac{p-2}{2}}$.
    
    \item Low-rank case: \\$\bm W_i^k = \bm I_q \otimes \left[\bm U_i^k\diag{\bm w_i}{\bm U_i^k}^\transp\pr{\bm Z_i^k {\bm Z_i^k}^\transp+\gamma\bm I}^{\frac{p-2}{2}}\right]$, \\ where $\bm Z^k_i = \bm U^k_i \diag{\bm \sigma(\bm Z^k_i)} {\bm V^k_i}^\transp$.
\end{itemize}

\item Find the updated solution $\bm x^{k+1}$ by solving the linear system:
\begin{align*}
& \hspace*{-1.5cm}\bm x^{k+1} = \\
& \hspace*{-1.5cm}\pr{\bm A^\transp\bm A + p\!\cdot\!\sg_{\bm n}^2\suml_{i=1}^\ell\bm G_i^\transp\bm W_i^k\bm G_i + \alpha\bm I}^{-1}\pr{\bm A^\transp\bm y +\alpha\bm x^k}.
\end{align*}

\item $k = k+1$.
\end{enumerate}
}
Return $\bm x^\ast = \bm x^k$\;
}
\Bwd{
\begin{enumerate}[wide, labelwidth=!, labelindent=0pt]
\footnotesize
\item Use $\bm x^\ast$ to compute $\bm W_i^\ast = \bm W_i^\ast\pr{\bm G, \bm x^*}$ following steps 1) \\and 2) in the \textbf{Forward Pass}. Then define the following auxiliary \\computational graph with parameters $\bm \theta$: \\ $\bm g\pr{\bm x^\ast, \bm \theta} = \pr{\bm A^\transp\bm A + p\!\cdot\!\sg_{\bm n}^2\suml_{i=1}^\ell\bm G_i^\transp\bm W_i^\ast\pr{\bm G, \bm x^\ast}\bm G_i}\bm x^\ast-\bm A^\transp\bm y$.
\item Compute $\bm v=\pr{\grad{\bm g}{\bm x^\ast}}^{-1}\bm\rho$ by solving the linear system \\$\grad{\bm g}{\bm x^\ast}\cdot\bm v = \bm \rho,$ where $\bm\rho =\grad{\mc L}{\bm x^\ast}$ and $\mc L$ is the
training loss function.
\item Obtain the gradient $\grad{\mc L}{\bm\theta}$ by computing the vector-jacobian \\product (VJP) $\grad{\bm g}{\bm\theta}\cdot\bm v$.
\item Use $\grad{\mc L}{\bm\theta}$ to update the network's parameters $\bm\theta$ or backpropagate further into their parents.
\vspace{-.2cm}
\end{enumerate}
}
 \caption{Forward and backward passes of LIRLS networks.}
  \label{alg:IBP}
\end{algorithm}

\subsection{Implementation details}\label{sec:LIRLS_implementation}
In this section we discuss the insights of how our proposed LIRLS networks could be implemented in practice. In our previous work~\cite{Lefkimmiatis2023} we have considered sparsity-enforcing weighted $\ell_p^p$ priors for grayscale problems and weighted $\mc S_p^p$ for color ones. In this work we extend our results by considering sparsity-enforcing priors for color images. We also utilise different forms of penalty functions, and we assess the difference in their performance on different image reconstruction tasks. In all our experiments we learn the parameters of the regularization operator $\bm G$, which we parameterize as a valid convolution layer that consists of 74 filters of size $3 \times 5\times 5$, meaning that these filters are mixing the color channels of the images. Further, depending on the particular LIRLS instance that we utilize, we either fix the values of the parameters $\bm w, p$ or we learn them during training. Hereafter, we will keep the same notation as in~\cite{Lefkimmiatis2023}, that is $\ell_{p}^{p, \bm w}$ and $\mc S_p^{p, \bm w}$ will refer to the networks that employ a learned sparse-promoting prior and a learned low-rank promoting prior, respectively. The different LIRLS instances that we consider are listed below:
\begin{enumerate}[leftmargin=*]
    \item $\ell_1$: fixed $p\!=\!1$, fixed $\bm w\!=\!\bm 1$, where $\bm 1$ is a vector of ones. This parametrization results in a convex problem.
    \item Weighted $\ell_1^{\bm w}$: fixed $p\!=\!1$,  weights $\bm w$ are computed by a weight prediction neural network (WPN). WPN accepts as inputs the features $\bm z_0\!=\!\hbm G \bm x_0$, as well as the noise standard deviation $\sigma_{\bm n}$. The vector $\bm x_0$ is the estimate obtained after 5 IRLS steps of the pretrained $\ell_1$ network, while $\hbm G$ is their learned regularization operator. For all studied problems, we use a lightweight RFDN architecture proposed by~\cite{Liu2020} to predict the weights. The number of parameters of WPN does not exceed $1$M. This parametrization also results in a convex problem.\label{item:1_wpn}
    \item $\ell_p^p$: $p$ is allowed to learn freely within the limits $p\!\in\![0.4, 0.9]$ (to avoid convergence instability issues), while the weights are kept fixed $\bm w\!=\!\bm 1$. This parametrization results in a non-convex problem.
    \item $\ell_p^{p,\bm w}$: learned $p\!\in\![0.4, 0.9]$, weights $\bm w$ are computed as described in item~\ref{item:1_wpn}, with the only difference being that both $\bm x_0$ and $\hbm G$ are now obtained from the pretrained $\ell_p^p$ network. This parametrization also results in a non-convex problem.
\end{enumerate}

An important point that we need to highlight is that if both operator $\bm G$ and weights $\bm w$ are allowed to be learned freely, then the interplay between them in the second term of the linear system matrix in Eq.~\eqref{eq:normal_eq} may result in an uncontrollable increase of one of them that will result to a vanishing behavior for the other. Indeed, if for example the amplitudes of $\bm w_i$ are changed by the multiplier $\eta$, and the amplitude of $\bm G_i$ is augmented with $1/\sqrt{\eta}$, such multiplier cancels out in the product $\bm G_i^\transp \bm W_i^k \bm G_i$. In order to avoid numerical instabilities related to this effect, during training of the weights prediction network for the $\ell_p^{p, \bm w}$ or $\ell_1^{\bm w}$ cases, the filters of the operator $\bm G$ are initialized from the corresponding $\ell_p^{p}$ or $\ell_1$ networks and are kept frozen throughout the whole training. While conducting our experiments we have tried to employ various normalization techniques for both $\bm G$ and WPN, including spectral normalization~\cite{Miyato2018}, activation functions and an explicit normalization by division. From the results of these unreported experiments we have concluded, that the joint learning of $\bm G$ and WPN makes the training unstable and does not improve the network's output quality.

We also note that the output of the LS layer, which corresponds to the solution of the normal equations in Eq.~\eqref{eq:normal_eq}, is computed by utilizing a preconditioned version of CG (PCG)~\cite{Hestenes1952}. This allows for an improved convergence of the linear solver. Finally, in all the reported cases, the constant $\delta$ related to the parameter $\alpha$ in Eq.~\eqref{eq:normal_eq} is set to $8e^{-4}$, while as initial solution for the linear solver we use the output of an independently trained fast FFT-based Wiener filter. 
\begin{table*}[t]
\centering
\caption{Comparisons on color image deblurring.}
\label{tab:deblur_color}
\tabcolsep=0.1cm
\begin{tabular}{ccccccccccccccccc}
\hline
 & & TVN & VTV & $\mc S_1$ & $\mc S_p^p$ & $\mc S_1^{\bm w}$ & $\mc S_p^{p,\bm w}$ & $\ell_1$ & $\ell_p^p$ & $\ell_1^{\bm w}$ & $\ell_p^{p, \bm w}$ & RED & IRCNN & FDN & DWDN & SVMAP \\ \hline
 Sun et al. & PSNR & 31.71  & 31.38 & 33.09 & 34.17 & 33.95 & 34.24 & 33.76 & 33.86 & \textbf{34.56} & 34.39 & 31.38 & 33.19 & 32.51 & 34.09 & 34.36\\
 \cite{Sun2013} & SSIM & 0.8506 & 0.8440 & 0.8995 & 0.9227 & 0.9179 & 0.9234 & 0.9122 & 0.9145 & \textbf{0.9287} & 0.9261 & 0.8233 & 0.9123 & 0.8857 & 0.9197 & 0.9249 \\ \hline
  BSD100 & PSNR & 29.72 & 29.49 & 30.17 & 30.68 & 31.22 & 30.94 & 30.66 & 30.65 & \textbf{31.61} & 31.39 & 29.30 & 30.03 & 29.68 & 30.95 & 31.53 \\
 \cite{Martin2001} & SSIM & 0.8340 & 0.8291 & 0.8477 & 0.8675 & 0.8785 & 0.8759 & 0.8638 & 0.8622 & \textbf{0.8893} & 0.8834 & 0.8155 & 0.8624 & 0.8419 & 0.8792 & 0.8861 \\ \hline
 Anger et al. & PSNR & 26.28 & 26.01 & 26.66 & 26.35 & \textbf{27.39} & 26.35 & 26.47 & 26.23 & 27.32 & 27.17 & 26.01 & 24.39 & 25.72 & 26.85 & 27.03 \\
 \cite{Anger2018} & SSIM & 0.8108 & 0.8019 & 0.8418 & 0.8435 & \textbf{0.8665} & 0.8432 & 0.8400 & 0.8303 & 0.8616 & 0.8626 & 0.8021 & 0.7564
 & 0.7839 & 0.8467 & 0.8558 \\ \hline
 & \# Params & 1 & 1 & 600 & 601 & 0.3M & 0.3M & 5550 & 5551 & 1.0M & 1.0M & 0.2M & 4.7M & 0.4M & 7M & 1.3M \\ \hline
\end{tabular}
\end{table*}
\begin{figure*}[b]
\centering
\setlength\extrarowheight{-5pt}
\begin{tabular}{@{} c @{} c @{} c @{} c @{} c @{} c @{} c @{} c @{} c @{}}
 % trim=left, bottom, right, top
 \includegraphics[width=.26267281105990786\linewidth]{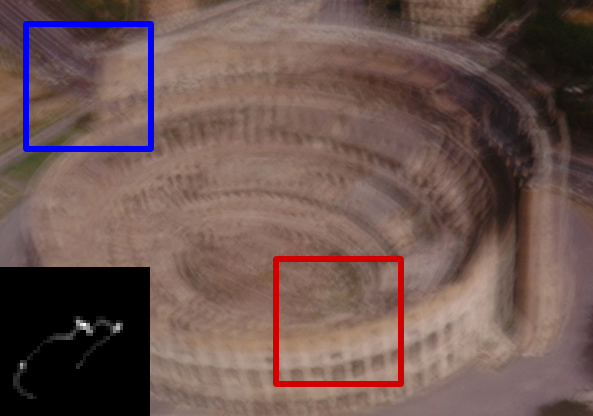} &
 \includegraphics[width=.09216589861751152\linewidth]{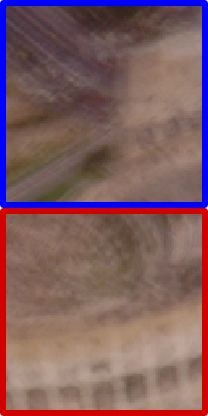} & 
 \includegraphics[width=.09216589861751152\linewidth]{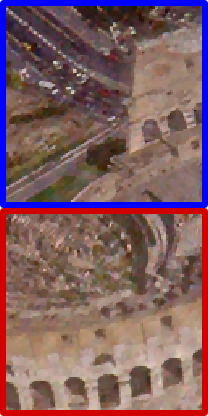} & 
 \includegraphics[width=.09216589861751152\linewidth]{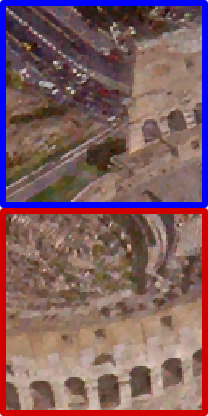} & 
 \includegraphics[width=.09216589861751152\linewidth]{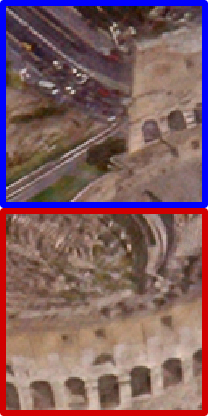} & 
 \includegraphics[width=.09216589861751152\linewidth]{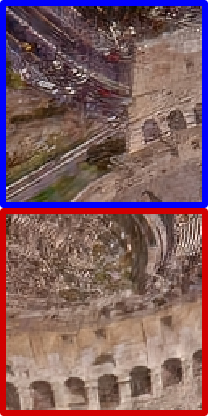} & 
 \includegraphics[width=.09216589861751152\linewidth]{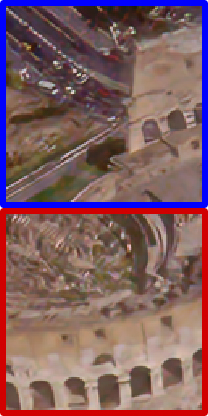} &
 \includegraphics[width=.09216589861751152\linewidth]{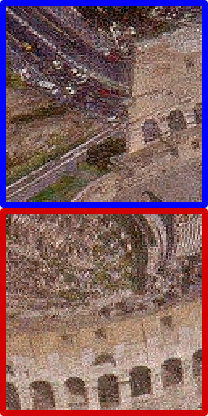} &
 \includegraphics[width=.09216589861751152\linewidth]{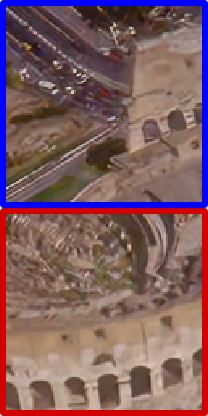} \\
 \footnotesize{Real blurred image} & \footnotesize{Input} & \footnotesize{VTV}& \footnotesize{TVN} & \footnotesize{RED} & \footnotesize{IRCNN} & \footnotesize{FDN} & \footnotesize{DWDN} & \footnotesize{SVMAP}\\

 \includegraphics[width=.26267281105990786\linewidth]{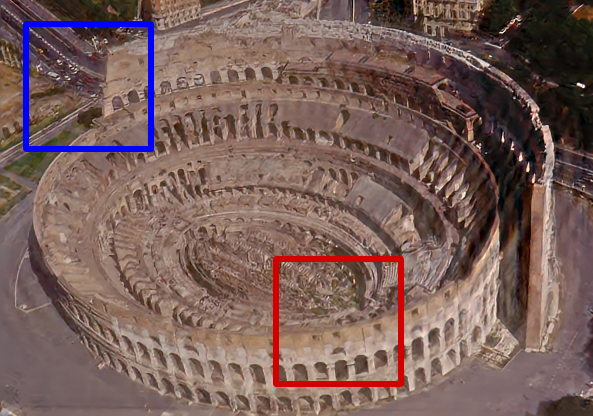} &
 \includegraphics[width=.09216589861751152\linewidth]{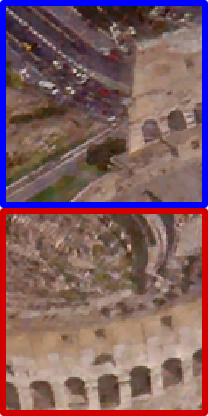} & 
 \includegraphics[width=.09216589861751152\linewidth]{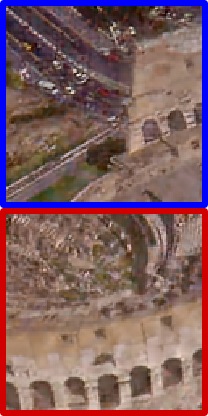} & 
 \includegraphics[width=.09216589861751152\linewidth]{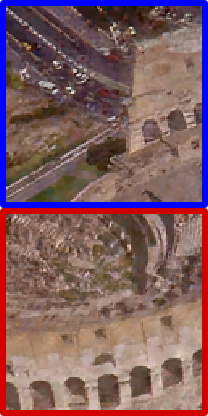} & 
 \includegraphics[width=.09216589861751152\linewidth]{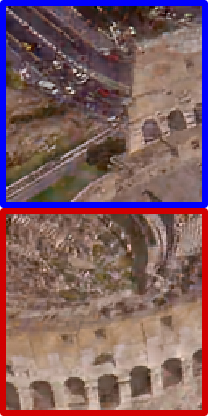} & 
 \includegraphics[width=.09216589861751152\linewidth]{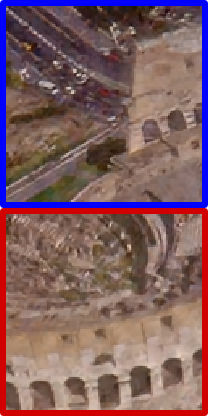} &
 \includegraphics[width=.09216589861751152\linewidth]{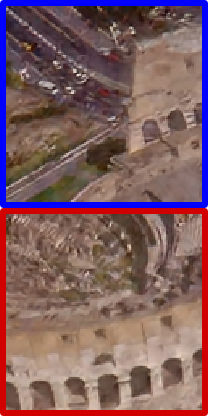} &
 \includegraphics[width=.09216589861751152\linewidth]{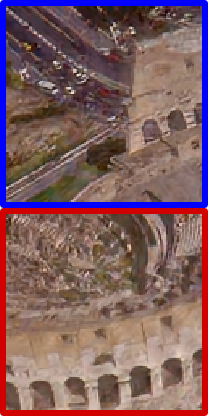} & 
 \includegraphics[width=.09216589861751152\linewidth]{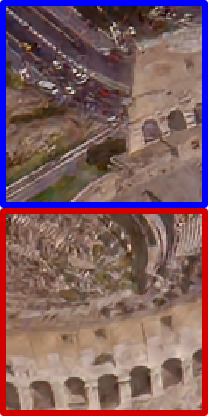}\\
 \footnotesize{Reconstructed image ($\ell_p^{\bm w}$ prior)} & \footnotesize{$\mc S_1$} & \footnotesize{$\mc S_p^p$}& \footnotesize{$\mc S_1^{\bm w}$} & \footnotesize{$\mc S_p^{\bm w}$} & \footnotesize{$\ell_1$} & \footnotesize{$\ell_p$} & \footnotesize{$\ell_1^{\bm w}$} & \footnotesize{$\ell_p^{\bm w}$}
\end{tabular}
   \caption{Visual comparisons among several methods on a real blurred color image (best viewed with x5 magnification). Image and blur kernel were taken from~\cite{Pan2016b}.}
   \label{fig:DeblurReal}
\end{figure*}
\vspace{-0.04\hsize}
\subsection{Training details}
We have enforced the convergence of LIRLS to a fixed point $\bm x^\ast$ according to the criterion:  $||\bm S^\ast\pr{\bm x^\ast,\bm\theta}\bm x^\ast-\bm A^\transp\bm y||_2 / ||\bm A^\transp\bm y||_2 < 1e^{-4}$, that has to be satisfied for 3 consecutive IRLS steps to perform an early exit. If this criterion is not satisfied, the forward pass is terminated after $400$ IRLS steps during training and $15$ steps during inference (In a few cases, which are explicitly mentioned in the text, we use a maximum of $30$ IRLS steps.). In the LS layers we use at most $150$ PCG iterations during training and perform an early exit when the relative tolerance of the unconditioned residual falls below the value of $1e^{-6}$, while during inference the maximum amount of CG iterations is reduced to $50$. When training LIRLS, in the backward pass, as shown in Eq.~\eqref{eq:backward_update}, we need to solve a linear problem whose system matrix matches the Hessian of the objective function \eqref{eq:objective}. This symmetric matrix is positive definite when $\mc J\pr{\bm x}$ is convex and indefinite otherwise. In the former case we utilize CG to solve the linear problem, while in the latter one we use the Minimal Residual Method (MINRES)~\cite{Paige1975}. In the backward stage we perform early exit if the relative tolerance of the residual is below $1e^{-2}$ and limit the maximum amount of iterations to $2000$. Throughout extensive experiments we have observed that these values ensure a stable training that requires a reasonable amount of time.

All our models are trained using the negative peak-to-signal-noise-ratio (PSNR) as loss function between the ground truth and the network's output. We use the Adam optimizer with the AMSGrad correction~\cite{Reddi2018} enabled and set a learning rate of $5e^{-3}$ for all models that do not involve a WPN and $1e^{-4}$ otherwise. The learning rate is decreased by a factor of 0.98 after each epoch. On average we set the batch size to 8 and train our models for 100 epochs, where each epoch consists of 500 batch passes. We have used double floating point precision everywhere during training and a single precision during inference.

\vspace{-0.05\hsize}
\section{Experiments}
Here we assess the performance of all LIRLS instances described in Sec.~\ref{sec:LIRLS_implementation} on three different color image reconstruction tasks, namely image deblurring, super-resolution and demosaicking. In all these cases the only difference in the objective function $\mc J\pr{\bm x}$ is the form of the degradation operator $\bm A$. Specifically, the operator $\bm A$ has one of the following forms: (\textbf{a}) low-pass valid convolution operator (\emph{deblurring}), (\textbf{b}) composition of a low-pass valid convolution operator and a decimation operator (\emph{super-resolution}), and (\textbf{c}) color filter array (CFA) operator (\emph{demosaicking}). In the deblurring and super-resolution tasks the low-pass valid convolution operator is applied independently on each of the three color channels.
\begin{figure*}[t]
\centering
\setlength\extrarowheight{-5pt}
\begin{tabular}{@{} c @{} c @{} c @{} c @{} c @{} c @{} c @{} c @{} c @{}}
 % trim=left, bottom, right, top
 \includegraphics[width=.3458128078817733\linewidth]{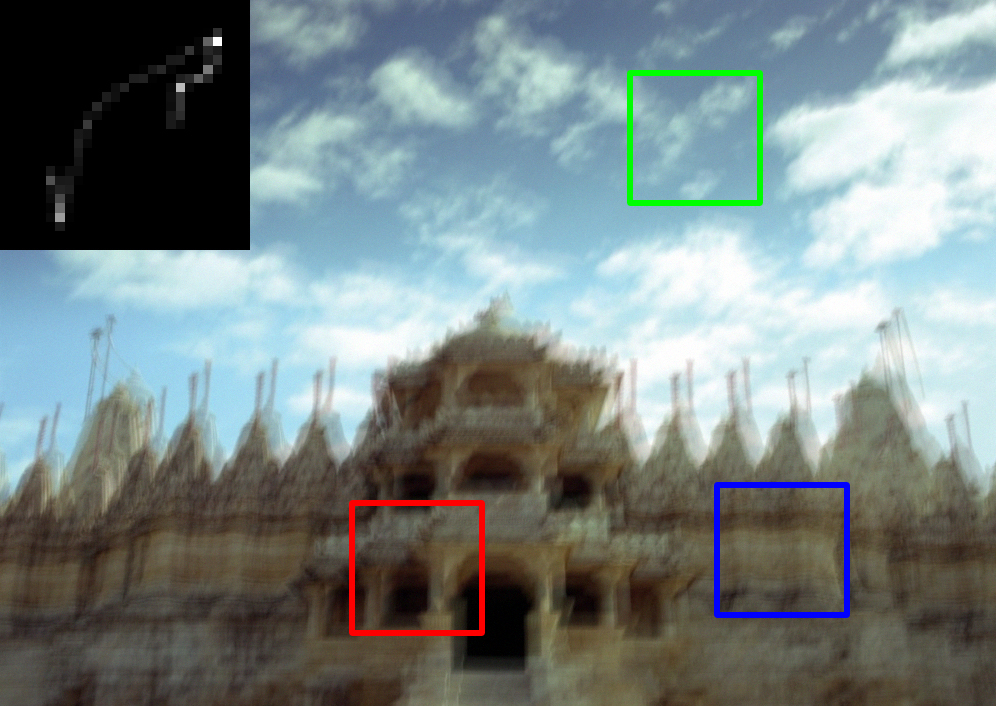} &
 \includegraphics[width=.08177339901477831\linewidth]{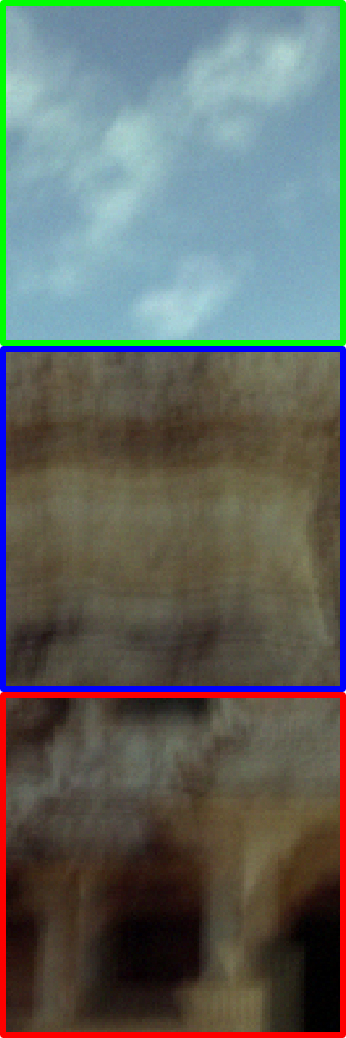} & 
 \includegraphics[width=.08177339901477831\linewidth]{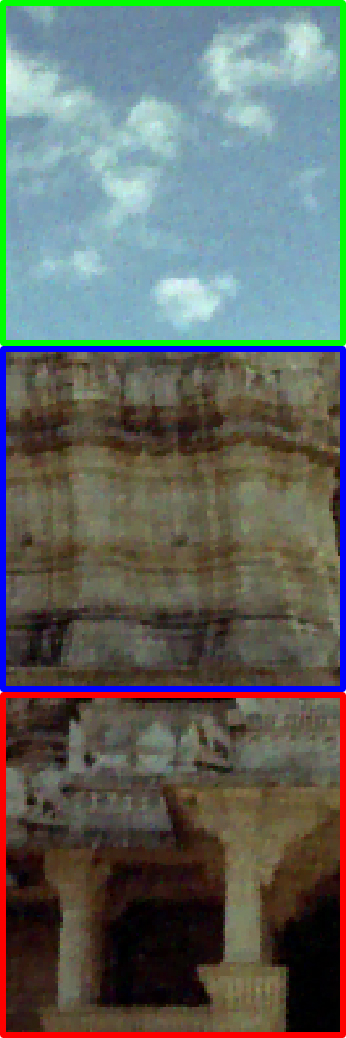} & 
 \includegraphics[width=.08177339901477831\linewidth]{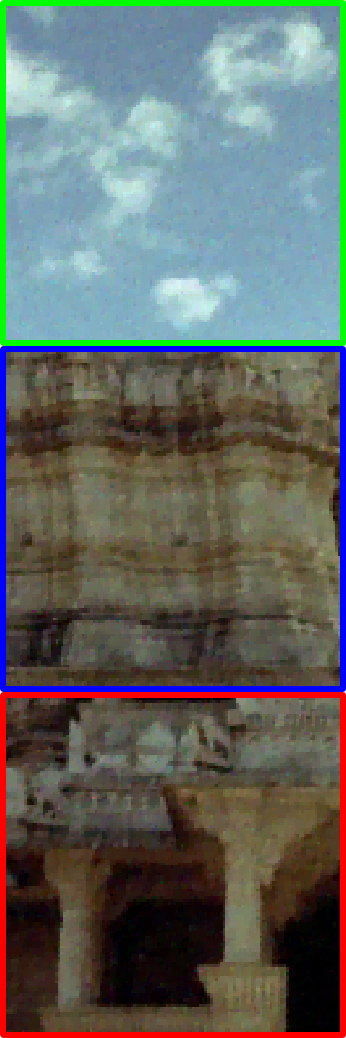} & 
 \includegraphics[width=.08177339901477831\linewidth]{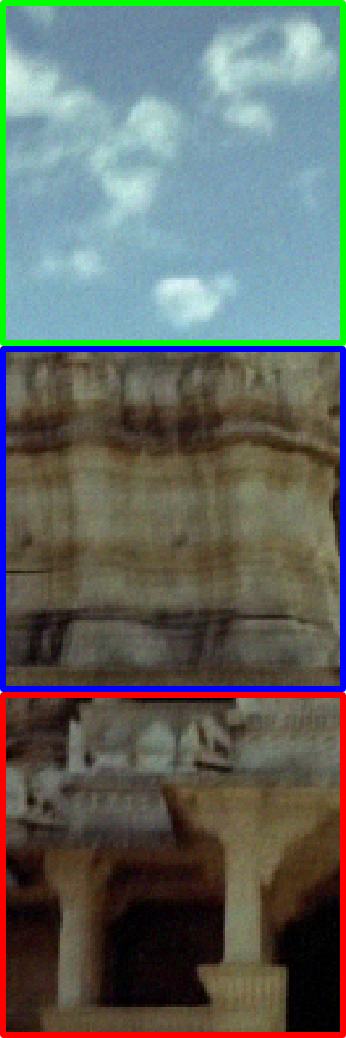} & 
 \includegraphics[width=.08177339901477831\linewidth]{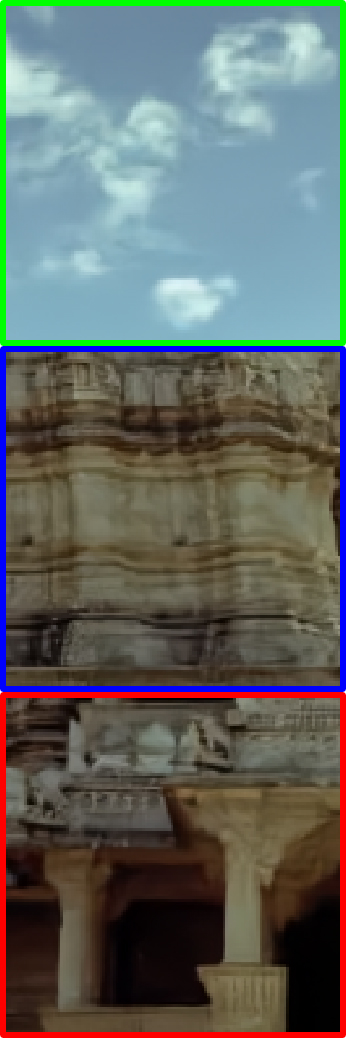} & 
 \includegraphics[width=.08177339901477831\linewidth]{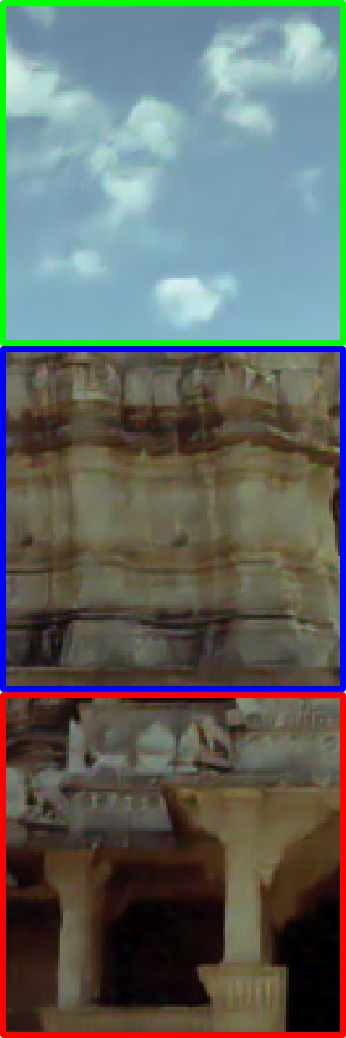} &
 \includegraphics[width=.08177339901477831\linewidth]{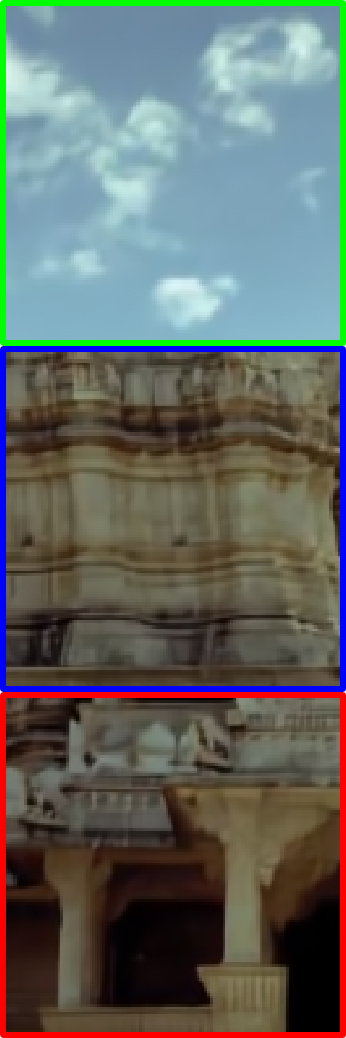} &
 \includegraphics[width=.08177339901477831\linewidth]{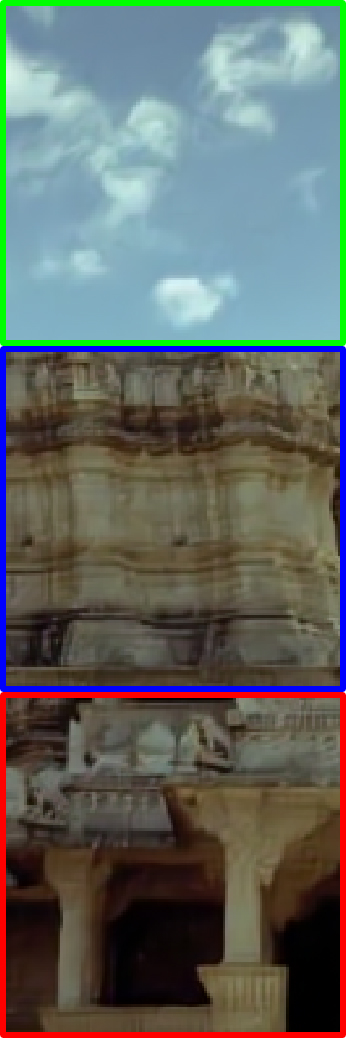} \\
 \footnotesize{Blurred image} & \footnotesize{Input} & \footnotesize{VTV}& \footnotesize{TVN} & \footnotesize{RED} & \footnotesize{IRCNN} & \footnotesize{FDN} & \footnotesize{DWDN} & \footnotesize{SVMAP}\\
 & \footnotesize{21.38} & \footnotesize{31.47}& \footnotesize{31.78} & \footnotesize{31.26} & \footnotesize{33.80} & \footnotesize{32.11} & \footnotesize{34.11} & \footnotesize{34.13}\\

 \includegraphics[width=.3458128078817733\linewidth]{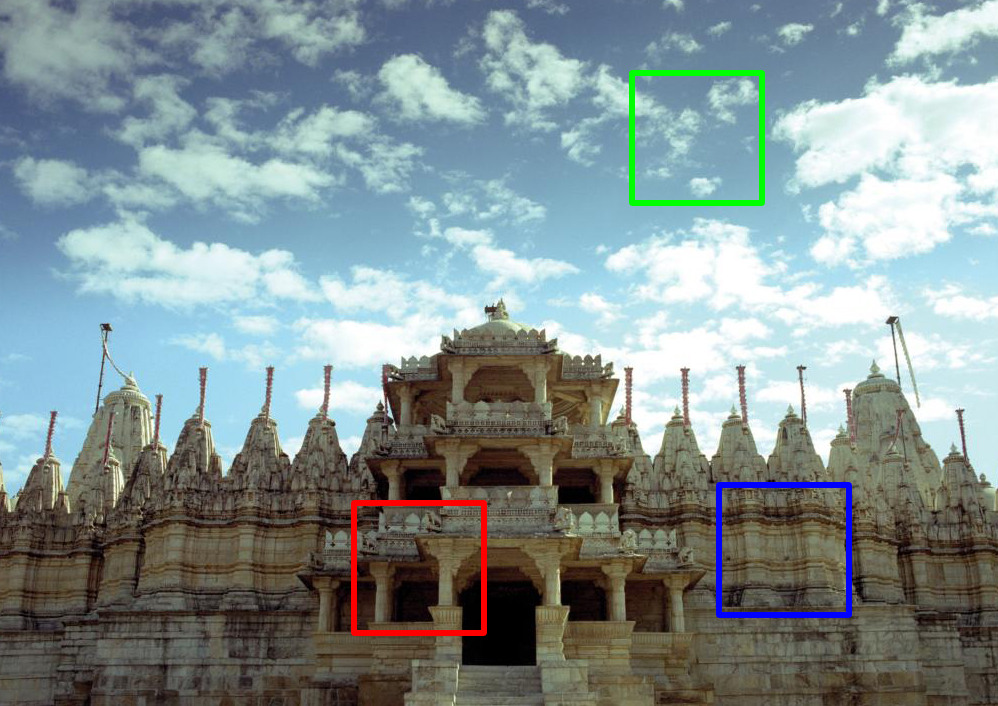} &
 \includegraphics[width=.08177339901477831\linewidth]{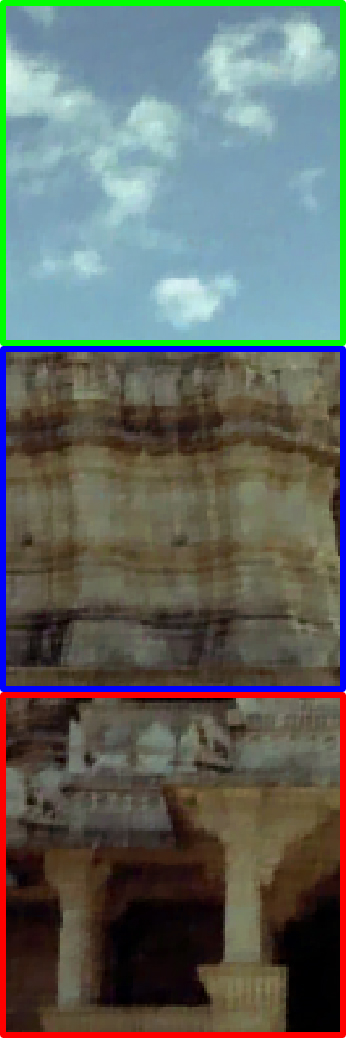} & 
 \includegraphics[width=.08177339901477831\linewidth]{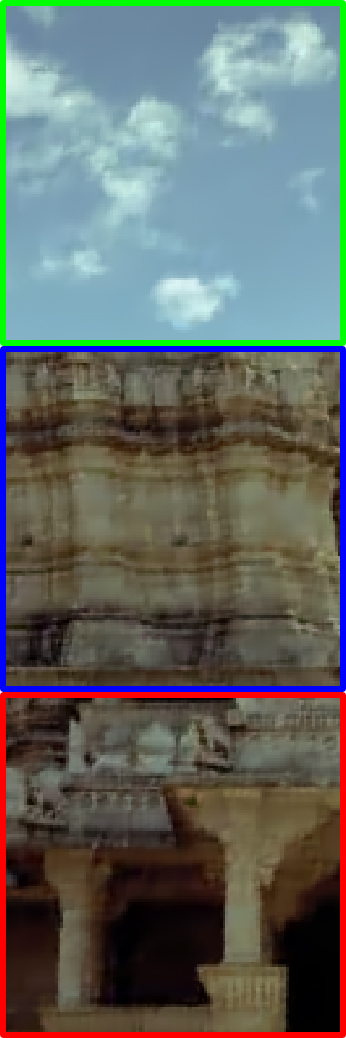} & 
 \includegraphics[width=.08177339901477831\linewidth]{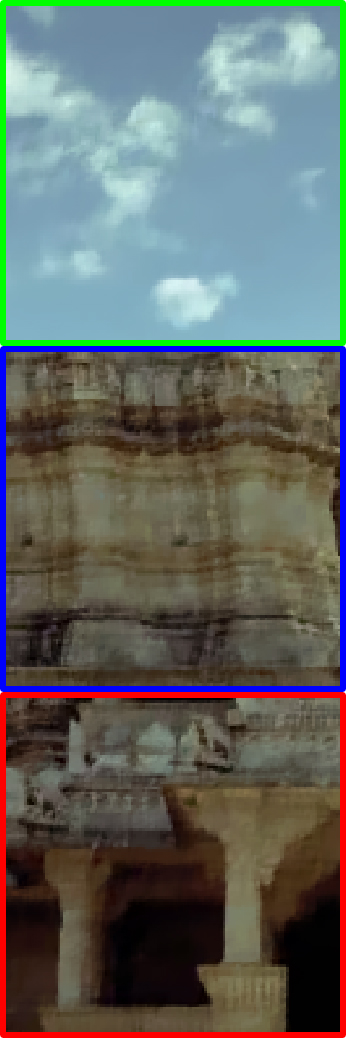} & 
 \includegraphics[width=.08177339901477831\linewidth]{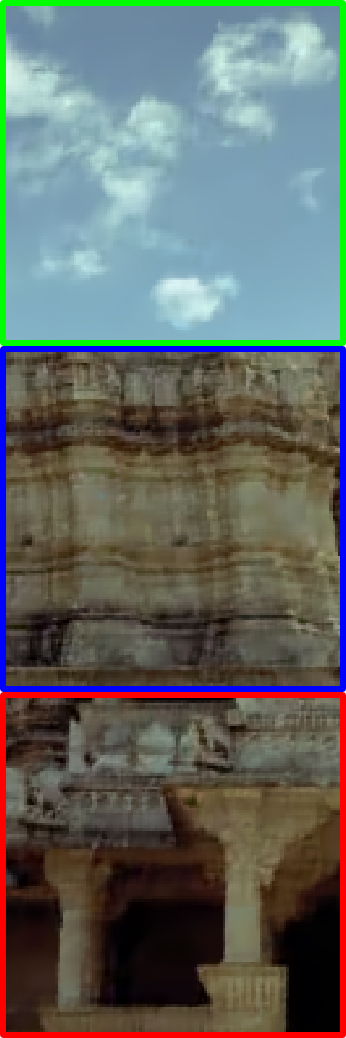} & 
 \includegraphics[width=.08177339901477831\linewidth]{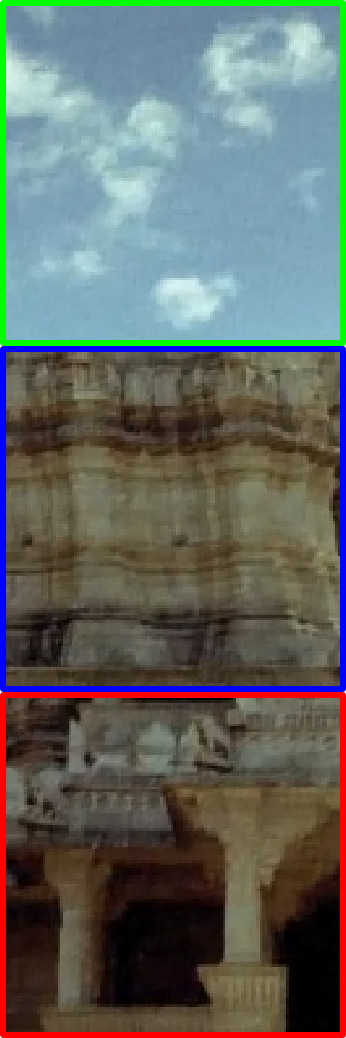} &
 \includegraphics[width=.08177339901477831\linewidth]{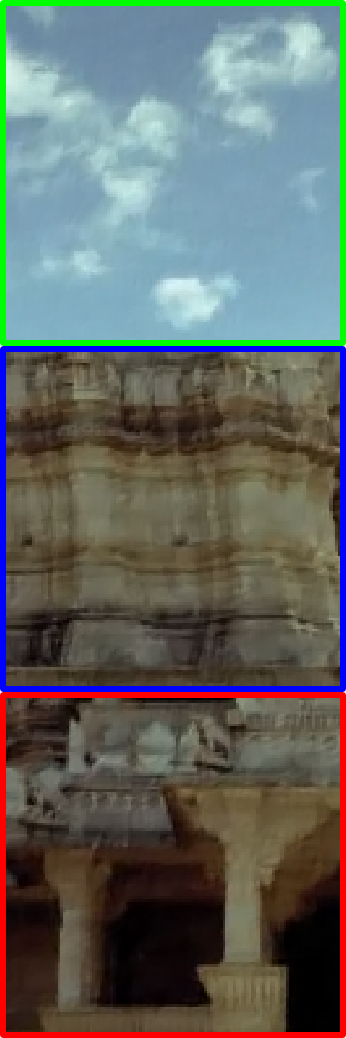} &
 \includegraphics[width=.08177339901477831\linewidth]{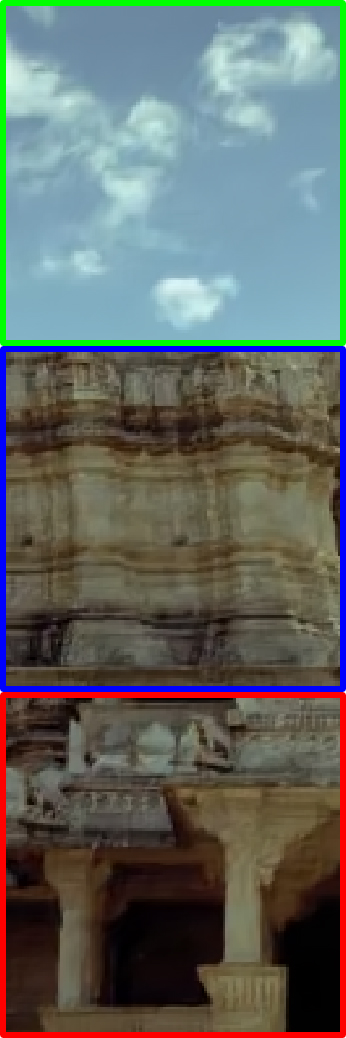} & 
 \includegraphics[width=.08177339901477831\linewidth]{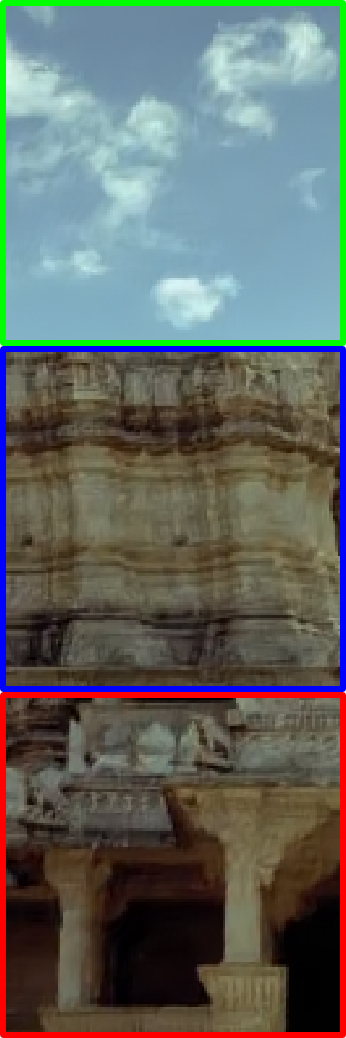}\\
 \footnotesize{Target image} & \footnotesize{$\mc S_1$} & \footnotesize{$\mc S_p^p$}& \footnotesize{$\mc S_1^{\bm w}$} & \footnotesize{$\mc S_p^{\bm w}$} & \footnotesize{$\ell_1$} & \footnotesize{$\ell_p$} & \footnotesize{$\ell_1^{\bm w}$} & \footnotesize{$\ell_p^{\bm w}$}\\
 & \footnotesize{33.34} & \footnotesize{34.52} & \footnotesize{34.11} & \footnotesize{34.53} & \footnotesize{34.05} & \footnotesize{34.06} & \footnotesize{\textbf{34.91}} & \footnotesize{34.68}
\end{tabular}
\caption{Visual comparisons among several methods on a synthetically blurred image from the~\cite{Sun2013} dataset (best viewed with x5 magnification). For each reconstructed image its PSNR value is provided in dB.}
\label{fig:DeblurSynthetic}
\end{figure*}
\begin{table*}[b]
\centering
\caption{Comparisons on color image super-resolution for BSD100RK dataset~\cite{Bell2019}.}
\label{tab:sr_color}
\tabcolsep=0.1cm
\begin{tabular}{cccccccccccccccccc}
\hline
Scale & Noise & & TVN & VTV & $\mc S_1$ & $\mc S_p^p$ & $\mc S_1^{\bm w}$ & $\mc S_p^{p,\bm w}$ & $\ell_1$& $\ell_p^p$& $\ell_1^{\bm{w}}$ & $\ell_p^{p, \bm{w}}$ & Bicubic & RED & IRCNN & ZSSR & USRNet \\ \hline
 \multirow{4}{*}{x2} & \multirow{2}{*}{$0\%$} & PSNR & 28.01 & 28.13 & 28.21 & 28.62 & 28.42 & 28.78 & 28.34 & 28.38 & \textbf{29.11} & 29.02 & 24.73 & 28.68 & 28.33 & 28.17 & 28.86 \\
 & & SSIM & 0.8035 & 0.8042 & 0.8088 & 0.8203 & 0.8212 & 0.8217 & 0.8113 & 0.8108 & 0.8321 & 0.8290 & 0.6661 & 0.8210 & 0.8331 & 0.8198 & \textbf{0.8359} \\
 & \multirow{2}{*}{$1\%$} & PSNR & 26.87 & 26.86 & 27.06 & 27.43 & 27.46 & 27.48 & 27.09 & 27.11 & 27.74 & 27.69 &  24.62 & 25.74 & 26.37 & 24.40 & \textbf{27.97} \\
 & & SSIM & 0.7404 & 0.7392 & 0.7469 & 0.7646 & 0.7640 & 0.7687 & 0.7520 & 0.7506 & 0.7790 & 0.7774 & 0.6486 & 0.6215 & 0.6843 & 0.5519 & \textbf{0.7924} \\ \hline
 \multirow{4}{*}{x3} & \multirow{2}{*}{$0\%$} & PSNR & 25.46 & 25.62 & 25.50 & 25.60 & 25.35 & 25.90 & 25.65 & 25.65 & \textbf{26.21} & 26.15 &  23.33 & 25.90 & 25.74 & 24.82 & 26.17  \\
 & & SSIM & 0.6812 & 0.6823 & 0.6828 & 0.6904 & 0.6877 & 0.6928 & 0.6861 & 0.6840 & 0.7048 & 0.7017 & 0.5833 & 0.7030 & 0.7092 & 0.6736 & \textbf{0.7203} \\
 & \multirow{2}{*}{$1\%$} & PSNR & 24.76 & 24.76 & 24.83 & 25.09 & 25.08 & 25.11 & 24.79 & 24.77 & 25.32 & 25.28 &  23.25 & 24.72 & 24.77 & 22.87 & \textbf{25.65} \\
 & & SSIM & 0.6322 & 0.6314 & 0.6348 & 0.6468 & 0.6452 & 0.6497 & 0.6368 & 0.6333 & 0.6575 & 0.6555 & 0.5702 & 0.6075 & 0.6203 & 0.4800 & \textbf{0.6857} \\ \hline
 \multirow{4}{*}{x4} & \multirow{2}{*}{$0\%$} & PSNR & 24.05 & 24.13 & 24.15 & 24.28 & 24.22 & 24.27 & 24.09 & 24.07 & \textbf{24.40} & 24.33 &  22.25 & 24.33 & 23.46 & 22.93 & 23.30  \\
 & & SSIM & 0.6018 & 0.6024 & 0.6045 & 0.6111 & 0.6079 & 0.6101 & 0.6047 & 0.6022 & 0.6127 & 0.6101 & 0.5305 & 0.6191 & 0.6073 & 0.5767 & \textbf{0.6238} \\
 & \multirow{2}{*}{$1\%$} & PSNR & 23.43 & 23.42 & 23.46 & \textbf{23.66} & 23.53 & \textbf{23.66} & 23.39 & 23.34 & 23.59 & 23.57 & 22.19 & 23.64 & 23.40 & 21.51 & 23.57 \\
 & & SSIM & 0.5655 & 0.5645 & 0.5666 & 0.5752 & 0.5698 & 0.5774 & 0.5670 & 0.5632 & 0.5718 & 0.5709 & 0.5210 & 0.5681 & 0.5643 & 0.4230 & \textbf{0.6028} \\ \hline
 & & \# Params & 1 & 1 & 600 & 601 & 0.3M & 0.3M & 5550 & 5551 & 1.0M & 1.0M & N/A & 0.2M & 4.7M & 0.2M & 17M \\ \hline
\end{tabular}
\end{table*}
\vspace{-0.08\hsize}
\subsection{Train and test data}\label{subsec:data}
To train our models for the mentioned recovery tasks, we use random crops of size $64 \times 64$ taken from the \emph{train} and \emph{val} subsets (400 images in total) of the BSD500 dataset provided by~\cite{Martin2001}. In order to train the deblurring networks we use synthetically created blur kernels varying in size from $13$ to $35$ pixels according to the procedure described by~\cite{Boracchi2012}, and for each kernel size we used a small subset of randomly synthesised kernels. For the super-resolution task we use scale factors $2$, $3$ and $4$ with a small set of $25\times 25$ kernels randomly synthesized using the algorithm provided by~\cite{Bell2019}. All our models are trained and evaluated on a range of noise levels that are typical in the literature for each considered problem. Based on this, we consider $\sigma_{\bm n}$ to be up to 1\% of the maximum image intensity for deblurring and super-resolution tasks, while for demosaicking we use a wider range of noise levels, with $\sigma_{\bm n}$ being up to 3\% in order to study the robustness of our method to the input noise.
\begin{figure*}[t]
\centering
\setlength\extrarowheight{-5pt}
\begin{tabular}{@{} c @{} c @{} c @{} c @{} c @{} c @{} c @{} c @{} c @{}}
 % trim=left, bottom, right, top
 \includegraphics[width=.30993456276026177\linewidth]{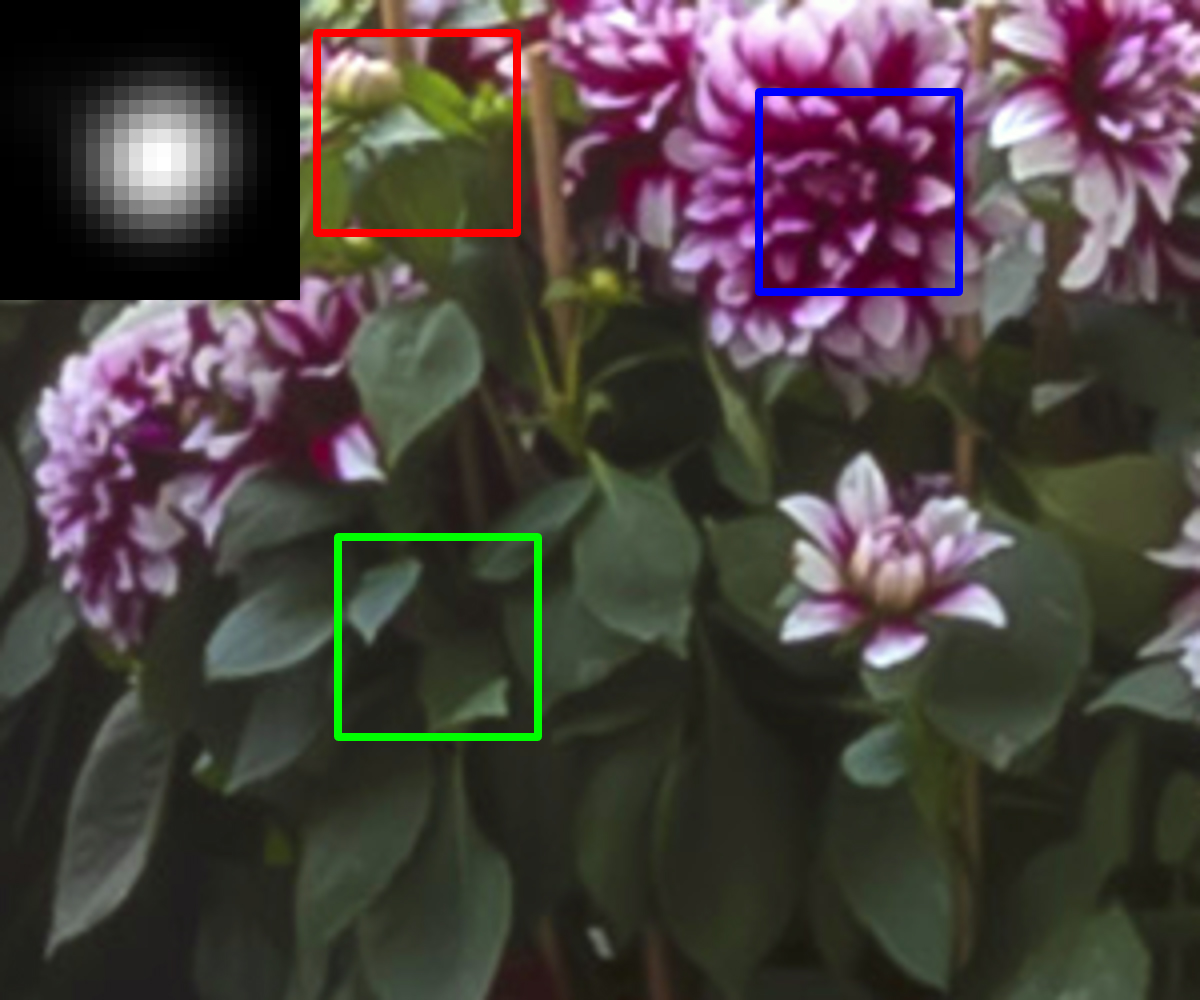} &
   & 
 \includegraphics[width=.08625817965496728\linewidth]{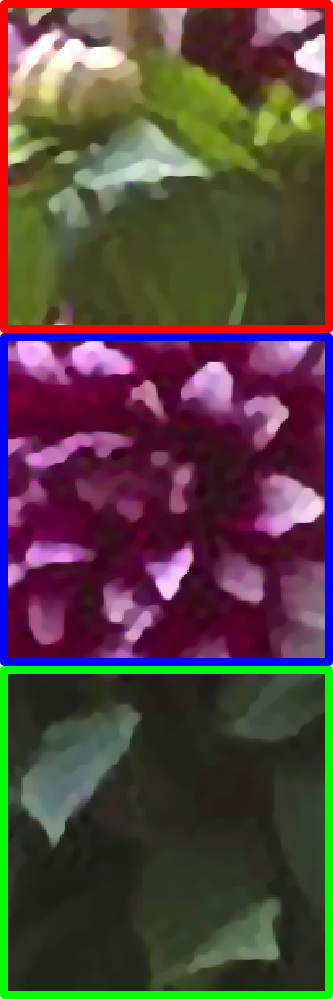} & 
 \includegraphics[width=.08625817965496728\linewidth]{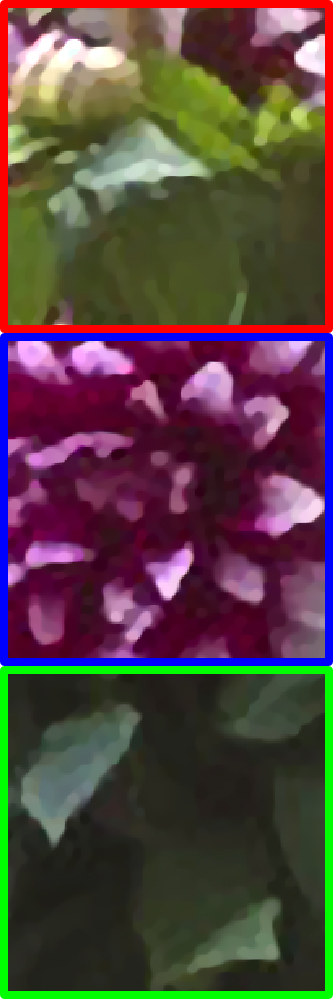} & 
  \includegraphics[width=.08625817965496728\linewidth]{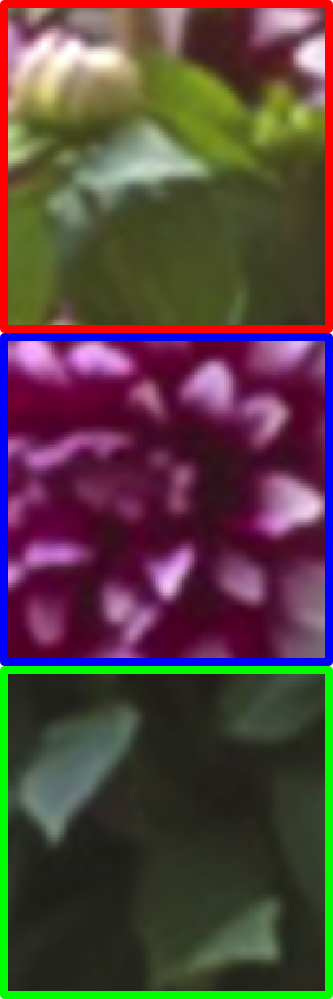} & 
 \includegraphics[width=.08625817965496728\linewidth]{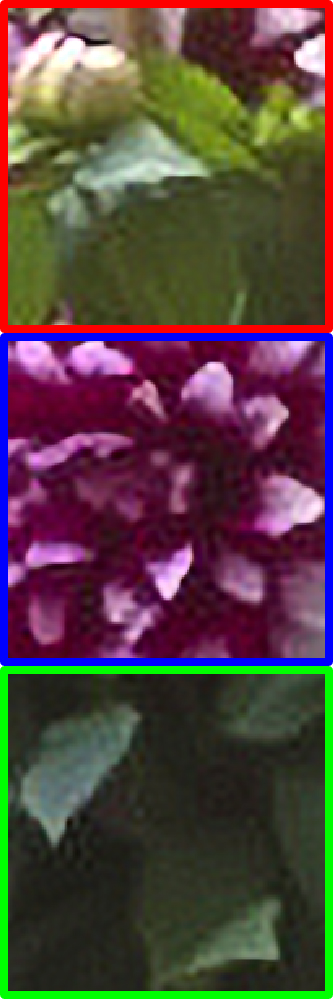} & 
 \includegraphics[width=.08625817965496728\linewidth]{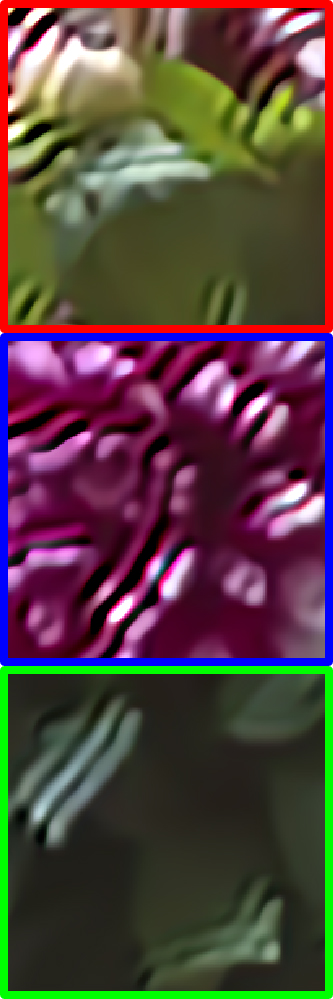} & 
 \includegraphics[width=.08625817965496728\linewidth]{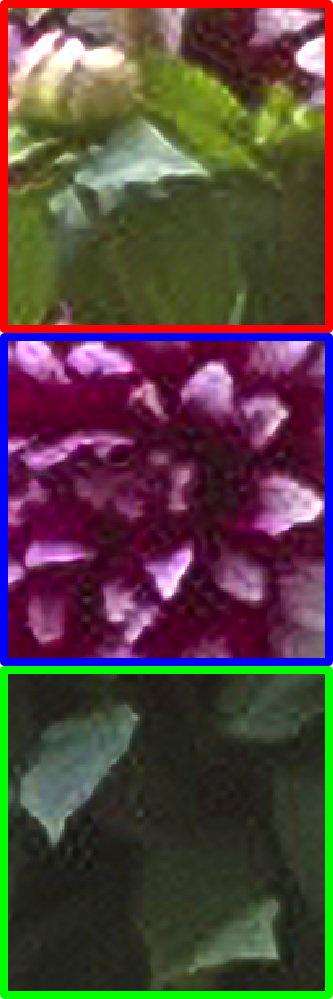} &
 \includegraphics[width=.08625817965496728\linewidth]{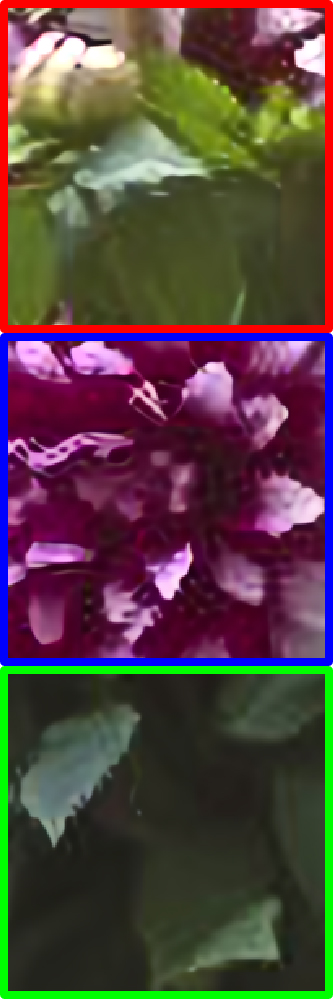} \\
 \footnotesize{Input (x4 bicubic upsampling)} & & \footnotesize{VTV}& \footnotesize{TVN} & \footnotesize{Bicubic} & \footnotesize{RED} & \footnotesize{IRCNN} & \footnotesize{ZSSR} & \footnotesize{USRNet} \\

 \includegraphics[width=.30993456276026177\linewidth]{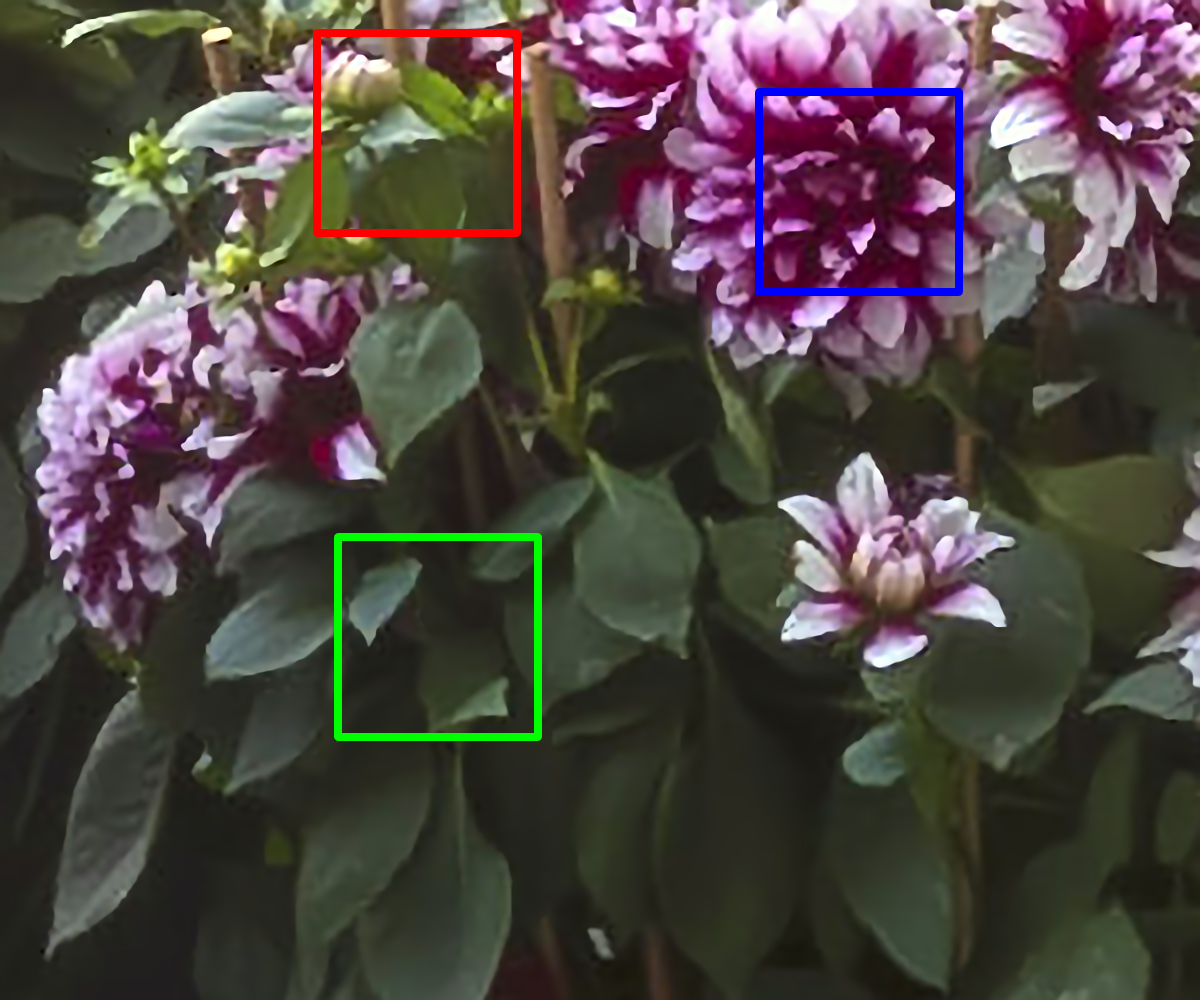} &
 \includegraphics[width=.08625817965496728\linewidth]{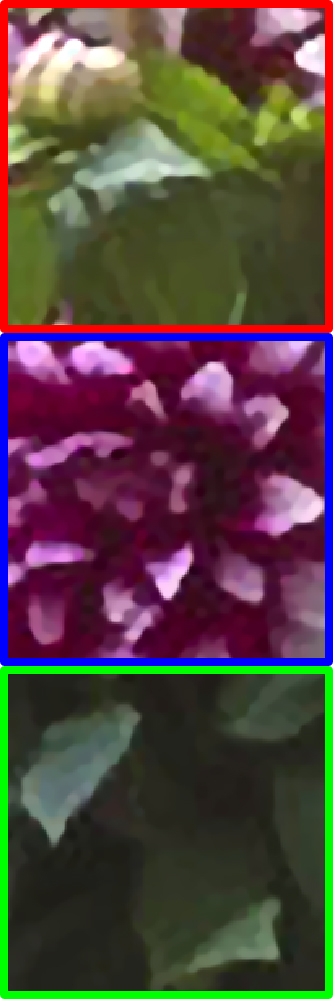} & 
 \includegraphics[width=.08625817965496728\linewidth]{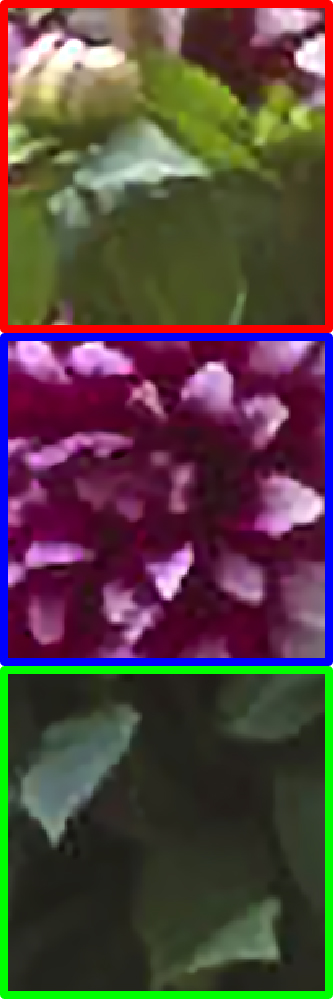} & 
 \includegraphics[width=.08625817965496728\linewidth]{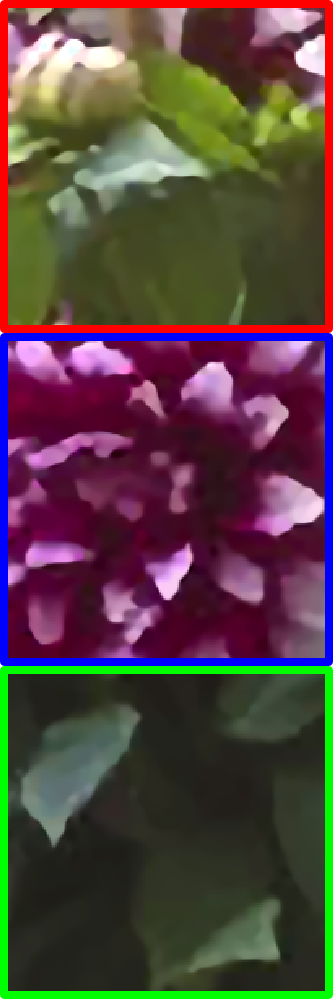} & 
 \includegraphics[width=.08625817965496728\linewidth]{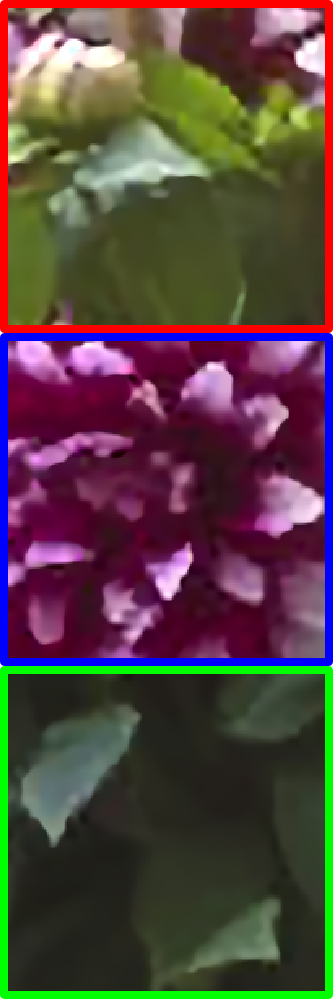} & 
 \includegraphics[width=.08625817965496728\linewidth]{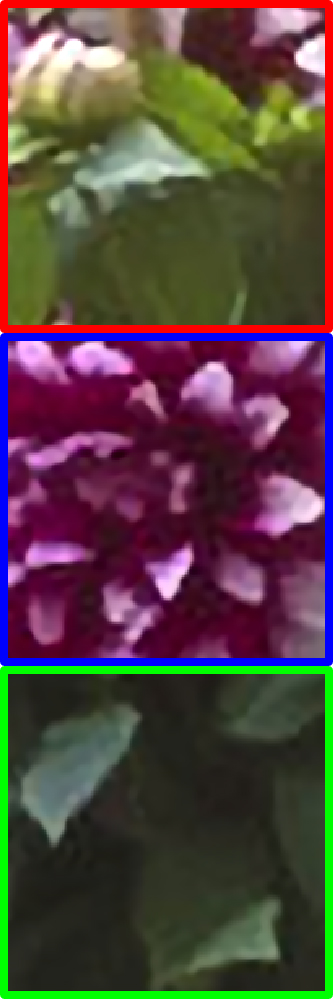} &
 \includegraphics[width=.08625817965496728\linewidth]{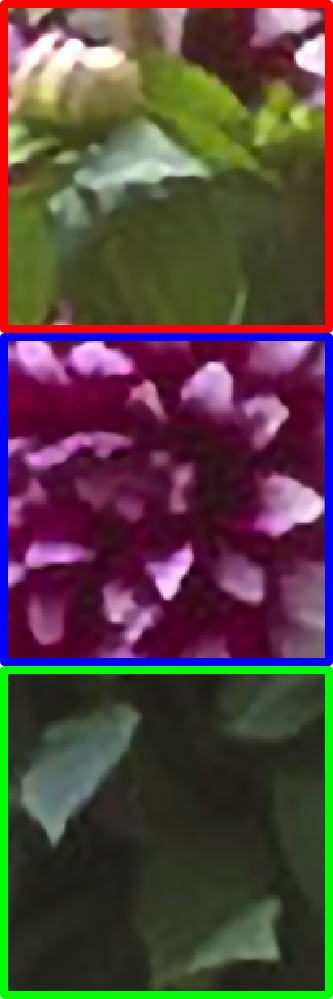} &
 \includegraphics[width=.08625817965496728\linewidth]{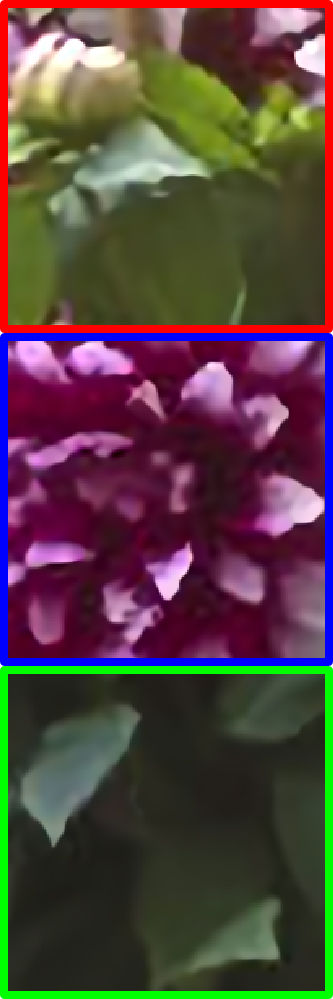} & 
 \includegraphics[width=.08625817965496728\linewidth]{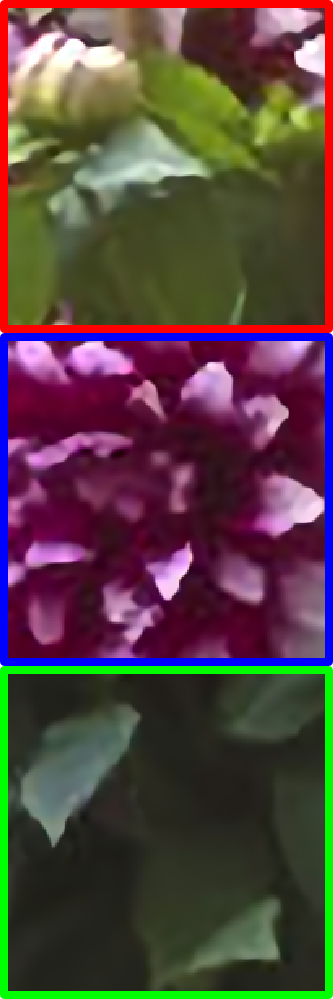}\\
 \footnotesize{Reconstructed image ($\ell_p^{\bm w}$ prior)} & \footnotesize{$\mc S_1$} & \footnotesize{$\mc S_p^p$}& \footnotesize{$\mc S_1^{\bm w}$} & \footnotesize{$\mc S_p^{\bm w}$} & \footnotesize{$\ell_1$} & \footnotesize{$\ell_p$} & \footnotesize{$\ell_1^{\bm w}$} & \footnotesize{$\ell_p^{\bm w}$}
\end{tabular}
\caption{Visual comparisons among several methods on a real x4 upsampled image from the RealSR dataset~\cite{Cai2019} (best viewed with x5 magnification). Downscaling kernel was estimated using the third-party method from~\cite{Bell2019}.}
\label{fig:SRRealX4}
\end{figure*}
For evaluation purposes we use diverse benchmarks, which have been proposed individually for each recovery task. In particular, for deblurring we use the 100 test images from the BSD500 dataset~\cite{Martin2001} with kernels randomly synthesised using the above described procedure and the addition of $1\%$ Gaussian noise. On top of that we use the common benchmark proposed by~\cite{Sun2013} (640 samples in total). Finally, in order to quantitatively evaluate the performance of our networks on images that resemble real samples, we use the benchmark of 8 images proposed in~\cite{Anger2018}. For super-resolution we use the BSD100RK dataset, which consists of 100 test images from the BSD500 dataset, and the degradation model proposed by~\cite{Bell2019}. For demosaicking we use the set of 18 images from the McMaster~\cite{Zhang2011} and a set of 24 images from the Kodak~\cite{Franzen1999} dataset. None of the data we use for evaluation purposes were presented to the networks during the training stage. The benchmarks we use for evaluation were specifically selected so as to contain a diverse set of images with relatively large resolutions. We should note that this strategy is not common for other general purpose methods, which typically report results on a limited set of small images, due to their need to manually fine-tune certain parameters. In our case all network parameters are learned via training and remain fixed during inference. In order to provide a fair comparison, we used a grid search to find the best set of parameters for the methods that do require manual tuning. Then we use these values fixed during the evaluation for the entire set of images per each benchmark. Finally, we note that for all TV-based regularizers whose results we report, the inference is performed with an IRLS strategy.

\subsection{Results}\label{sec:Results}
\begin{figure*}[t]
\centering
\setlength\extrarowheight{-5pt}
\begin{tabular}{@{} c @{} c @{} c @{} c @{} c @{} c @{} c @{} c @{} c @{}}
 % trim=left, bottom, right, top
 \includegraphics[width=.36318407960199006\linewidth]{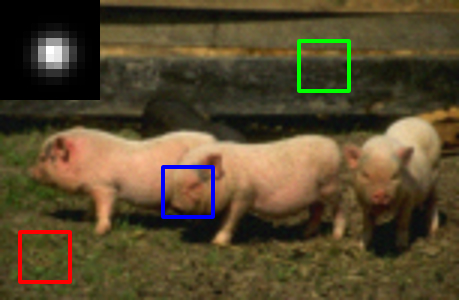} &
  \includegraphics[width=.07960199004975126\linewidth]{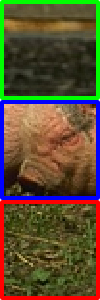} & 
 \includegraphics[width=.07960199004975126\linewidth]{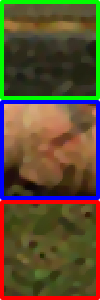} & 
 \includegraphics[width=.07960199004975126\linewidth]{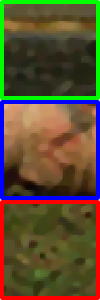} & 
  \includegraphics[width=.07960199004975126\linewidth]{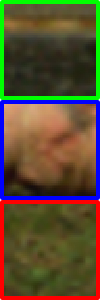} & 
 \includegraphics[width=.07960199004975126\linewidth]{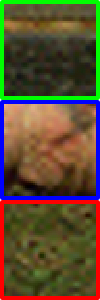} & 
 \includegraphics[width=.07960199004975126\linewidth]{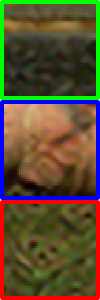} & 
 \includegraphics[width=.07960199004975126\linewidth]{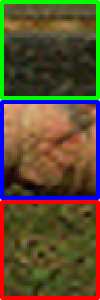} &
 \includegraphics[width=.07960199004975126\linewidth]{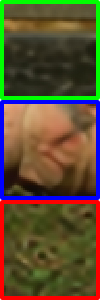} \\
 \footnotesize{Input (x3 bicubic upsampling)} & \footnotesize{Target} & \footnotesize{VTV}& \footnotesize{TVN} & \footnotesize{Bicubic} & \footnotesize{RED} & \footnotesize{IRCNN} & \footnotesize{ZSSR} & \footnotesize{USRNet} \\
 & & \footnotesize{26.26}& \footnotesize{26.30} & \footnotesize{25.21} & \footnotesize{26.15} & \footnotesize{26.14} & \footnotesize{25.58} & \footnotesize{\textbf{26.76}}\\

 \includegraphics[width=.36318407960199006\linewidth]{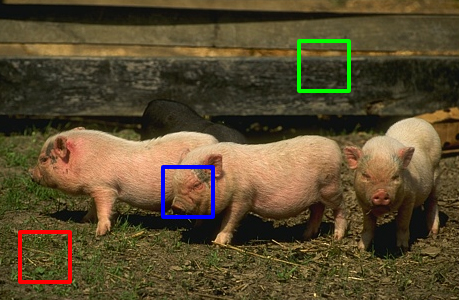} &
 \includegraphics[width=.07960199004975126\linewidth]{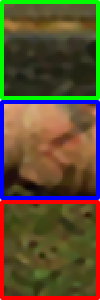} & 
 \includegraphics[width=.07960199004975126\linewidth]{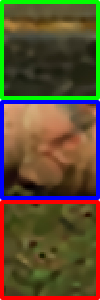} & 
 \includegraphics[width=.07960199004975126\linewidth]{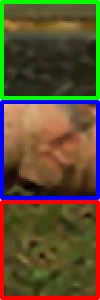} & 
 \includegraphics[width=.07960199004975126\linewidth]{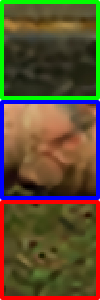} & 
 \includegraphics[width=.07960199004975126\linewidth]{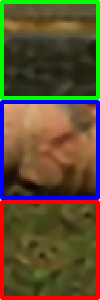} &
 \includegraphics[width=.07960199004975126\linewidth]{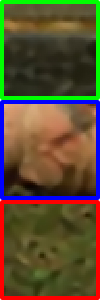} &
 \includegraphics[width=.07960199004975126\linewidth]{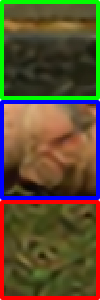} & 
 \includegraphics[width=.07960199004975126\linewidth]{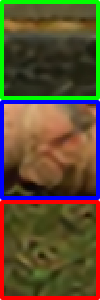}\\
 \footnotesize{Target image} & \footnotesize{$\mc S_1$} & \footnotesize{$\mc S_p^p$}& \footnotesize{$\mc S_1^{\bm w}$} & \footnotesize{$\mc S_p^{\bm w}$} & \footnotesize{$\ell_1$} & \footnotesize{$\ell_p$} & \footnotesize{$\ell_1^{\bm w}$} & \footnotesize{$\ell_p^{\bm w}$}\\
 & \footnotesize{26.44} & \footnotesize{26.52} & \footnotesize{26.53} & \footnotesize{26.56} & \footnotesize{26.45} & \footnotesize{26.44} & \footnotesize{26.60} & \footnotesize{26.62}
\end{tabular}
\caption{Visual comparisons among several methods on a x3 synthetically downsampled image from the BSD100RK dataset with 1\% noise (best viewed with x5 magnification). For each reconstructed image its PSNR value is provided in dB.}
\label{fig:SRSynthetic}
\end{figure*}
\begin{table*}[b]
\caption{Comparisons on image demosaicking.}
\label{tab:demosaic_color}
\tabcolsep=0.1cm
\centering
\begin{tabular}{cccccccccccccccc}
\hline
 Dataset & Noise & & TVN & VTV & $\mc S_1$ & $\mc S_p^p$ & $\mc S_1^{\bm w}$ & $\mc S_p^{p, \bm w}$ & $\ell_1$& $\ell_p^p$& $\ell_1^{\bm{w}}$ & $\ell_p^{p, \bm{w}}$ & Bilinear & RED & IRCNN \\ \hline
 & \multirow{2}{*}{$0\%$} & PSNR & 30.83 & 30.18 & 34.69 & 36.29 & 36.52 & 36.35 & 40.98 & 40.53 & \textbf{42.09} & $41.02^*$ & 29.12 & 36.48 & 40.48 \\
 & & SSIM & 0.9137 & 0.8939 & 0.9647 & 0.9730 & 0.9765 & 0.9728 & 0.9863 & 0.9828 & \textbf{0.9874} & $0.9855^*$ & 0.8835 & 0.9582 & 0.9811 \\
 & \multirow{2}{*}{$1\%$} & PSNR & 30.65 & 29.92 & 33.86 & 35.27 & 35.27 & 35.05 & 38.28 & 38.29 & \textbf{39.31} & $38.58^*$ & 28.88 & 35.68 & 38.17 \\
 Kodak & & SSIM & 0.8949 & 0.8753 & 0.9442 & 0.9559 & 0.9590 & 0.9550 & 0.9648 & 0.9660 & \textbf{0.9711} & $0.9687^*$ & 0.8620 & 0.9422 & 0.9609 \\
 \cite{Franzen1999} & \multirow{2}{*}{$2\%$} & PSNR & 30.01 & 29.31 & 32.36 & 33.34 & 33.61 & 33.52 & 35.60 & 35.94 & \textbf{36.70} & $36.09^*$ & 28.27 & 33.95 & 35.78 \\
 & & SSIM & 0.8593 & 0.8339 & 0.9102 & 0.9152 & 0.9308 & 0.9269 & 0.9366 & 0.9426 & \textbf{0.9490} & $0.9446^*$ & 0.8080 & 0.9003 & 0.9372 \\
 & \multirow{2}{*}{$3\%$} & PSNR & 29.37 & 28.59 & 30.58 & 31.23 & 32.26 & 32.19 & 33.50 & 33.78 & \textbf{34.88} & $34.29^*$ & 27.44 & 32.07 & 34.02 \\
 & & SSIM & 0.8269 & 0.7871 & 0.8614 & 0.8540 & 0.9013 & 0.8941 & 0.9085 & 0.9047 & \textbf{0.9274} & $0.9199^*$ & 0.7389 & 0.8441 & 0.9048 \\ \hline
 & \multirow{2}{*}{$0\%$} & PSNR & 33.33 & 32.64 & 36.27 & \textbf{37.66} & 36.60 & 37.62 & 35.85 & 36.69 & 37.41 & 36.68 & 32.27 & 34.53 & 37.52 \\
 & & SSIM & 0.9377 & 0.9285 & 0.9605 & \textbf{0.9649} & 0.9606 & 0.9632 & 0.9492 & 0.9570 & 0.9575 & 0.9570 & 0.9259 & 0.9338 & 0.9612 \\
 & \multirow{2}{*}{$1\%$} & PSNR & 33.32 & 32.16 & 35.14 & \textbf{36.43} & 35.49 & 36.37 & 34.83 & 35.65 & 36.30 & 35.70 & 31.76 & 34.40 & 36.12 \\
McMaster & & SSIM & 0.9198 & 0.9076 & 0.9416 & \textbf{0.9498} & 0.9451 & 0.9490 & 0.9294 & 0.9405 & 0.9441 & 0.9421 & 0.9014 & 0.9208 & 0.9403 \\
\cite{Zhang2011} & \multirow{2}{*}{$2\%$} & PSNR & 32.21 & 31.16 & 33.37 & 34.55 & 33.95 & 34.80 & 33.52 & 34.26 & \textbf{34.92} & 34.36 & 30.64 & 33.08 & 34.42 \\
 & & SSIM & 0.8843 & 0.8638 & 0.9151 & 0.9224 & 0.9226 & \textbf{0.9281} & 0.9075 & 0.9210 & 0.9269 & 0.9221 & 0.8443 & 0.8821 & 0.9178 \\
 & \multirow{2}{*}{$3\%$} & PSNR & 31.19 & 30.13 & 31.39 & 32.44 & 32.60 & 33.34 & 32.24 & 32.76 & \textbf{33.72} & 33.14 & 29.34 & 31.59 & 33.23 \\
 & & SSIM & 0.8539 & 0.8164 & 0.8808 & 0.8835 & 0.8997 & 0.9050 & 0.8900 & 0.8960 & \textbf{0.9102} & 0.9022 & 0.7745 & 0.8308 & 0.8931 \\ \hline
  & & \# Params & 1 & 1 & 600 & 601 & 0.3M & 0.3M & 5550 & 5551 & 1.0M & 1.0M & N/A & 0.2M & 4.7M \\ \hline
\end{tabular}
\end{table*}

\textbf{Deblurring.} In Table~\ref{tab:deblur_color} we report the average results in terms of PSNR and structure-similarity index measure (SSIM). We compare the previously reported LIRLS instances which utilize different low-rank promoting priors, our newly proposed and trained sparsity-promoting models and few competing methods. Regarding the competing methods we report the vector-valued TV (VTV)~\cite{Blomgren1998}, Nuclear TV (TVN)~\cite{Lefkimmiatis2015J}, RED~\cite{Romano2017}, IRCNN~\cite{Zhang2017b}, FDN~\cite{Kruse2017}, DWDN~\cite{Dong2020} and SVMAP~\cite{Dong2021}. From these results we observe that our newly proposed LIRLS models lead to very competitive performance, despite the relatively small number of learned parameters. In fact, it is noteworthy that our $\ell_1^{\bm w}$ variant obtains the best results on BSD100 and Sun \etal datasets when compared to all other methods, including SVMAP and DWDN, which are the current sota methods, and IRCNN that involves 4.7M parameters in total (for its 25 denoising networks). For the Anger \etal dataset, the $\ell_1^{\bm w}$ model also outperforms all the third-party competitors, however the best performance on this data is delivered by the $\mc S_1^{\bm w}$ model. From the same table we can also observe, that the weighted sparsity-promoting priors perform better than their low-rank enforcing counterparts. However, without the weight prediction network we can not observe a stable relation, since $\mc S_1$ prior performs worse by a large margin than the corresponding $\ell_1$ one, while $\mc S_p^p$ outperforms the $\ell_p^p$. Another conclusion we can draw is that in low-rank priors the norm order $p$ plays a more important role than in sparsity-enforcing ones, since the differences in performance between $\mc S_1$ and $\mc S_p^p$ models as well as $\mc S_1^{\bm w}$ and $\mc S_p^{p, \bm w}$ ones are significantly larger than those between $\ell_1$ and $\ell_p^p$ or $\ell_1^{\bm w}$ and $\ell_p^{p, \bm w}$. We also provide a visual example of the performance of our networks in Fig.~\ref{fig:DeblurReal} for a real image deblurring example and in Fig.~\ref{fig:DeblurSynthetic} for a synthetic one. From both examples it is clear, that our learned sparsity-promoting $\ell_p^{\bm w}$ and $\ell_1^{\bm w}$ regularization is able to recover fine details of the underlying image with a very high fidelity and without amplifying the input's noise. It is also worth noting that our models work well even when the estimated blur kernel is used instead of the ground truth one. This can be attributed to the fact that our proposed networks involve significantly less parameters than alternative networks and thus are less vulnerable to overfitting on the training data.

\textbf{Super-resolution.} Similarly to the image deblurring problem, we provide comparisons among competing methods for color images in Table~\ref{tab:sr_color}. For these comparisons we consider the RED and IRCNN methods as in the case of deblurring and we further include results from bicubic upsampling, ZSSR~\cite{Shocher2018} and USRNet~\cite{Zhang2020} networks, which has either similar or significantly larger number of learned parameters compared to our networks. Unlike RED and IRCNN, ZSSR and USRNet are specifically designed to deal with the problem of non-blind super-resolution and involve 0.2M and 17M of learned parameters, respectively. From the reported results we can see that for the color super-resolution the performance we obtain with our LIRLS instances is very competitive and it is in some cases only slightly inferior than that of the specialized USRNet networks. From the reported super-resolution experiments we can draw similar conclusions as with the deblurring case, meaning that the $\ell_1^{\bm w}$ model on average performs the best, the value of $p$ plays more crucial role for the low-rank priors than for the sparsity-enforcing ones, while the weights $\bm w$ act in the opposite way. For a visual assessment of the reconstruction quality we refer to Fig.~\ref{fig:SRRealX4} for a real example and to Fig.~\ref{fig:SRSynthetic} for a synthetic one. From these examples we clearly see that for the synthetic data our models perform almost the same way as the state-of-the-art USRNet method. At the same time for the real case scenarios, when an accurate downscale kernel is not available, LIRLS clearly outperforms all of the competitors by a large margin, since it restores a better level of details while keeping low the amount of ringing artifacts.
\begin{figure*}[t]
\centering
\setlength\extrarowheight{-5pt}
\begin{tabular}{@{} c @{} c @{} c @{} c @{} c @{} c @{} c @{} c @{}}
 % trim=left, bottom, right, top
 \includegraphics[width=.39130405009474456\linewidth]{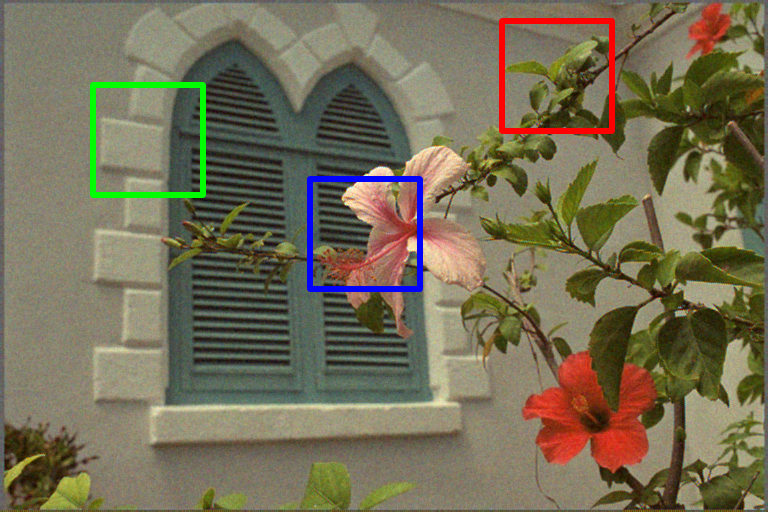} & 
 \includegraphics[width=.08695656427217936\linewidth]{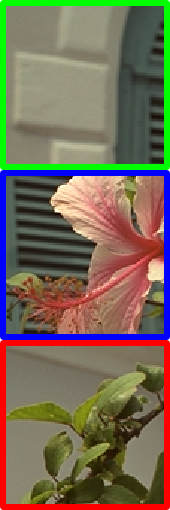} & 
 \includegraphics[width=.08695656427217936\linewidth]{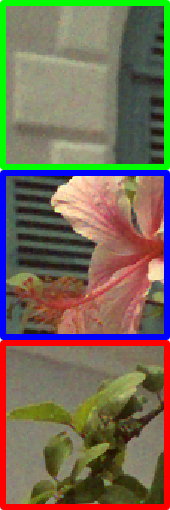} & 
 \includegraphics[width=.08695656427217936\linewidth]{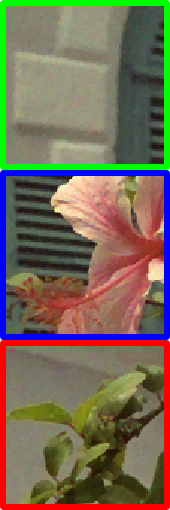} & 
 \includegraphics[width=.08695656427217936\linewidth]{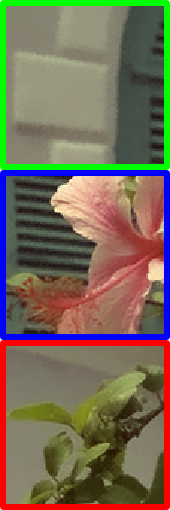} &
 \includegraphics[width=.08695656427217936\linewidth]{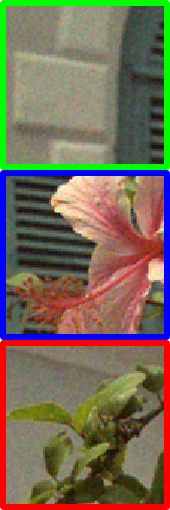} &
 \includegraphics[width=.08695656427217936\linewidth]
 {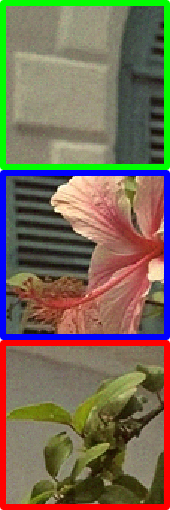} & 
 \includegraphics[width=.08695656427217936\linewidth]{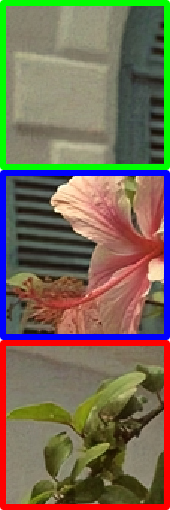} \\
 \multirow{2}{*}{\footnotesize{Input (bilinearly demosaicked)}} & \footnotesize{Target} & \footnotesize{VTV} & \footnotesize{TVN} & \footnotesize{$\mc S_1$} & \footnotesize{Bilinear} & \footnotesize{RED} & \footnotesize{IRCNN}\\
 & & \footnotesize{30.57} & \footnotesize{31.42} & \footnotesize{32.07} & \footnotesize{29.20} & \footnotesize{33.05} & \footnotesize{35.50} \\

  \includegraphics[width=.39130405009474456\linewidth]{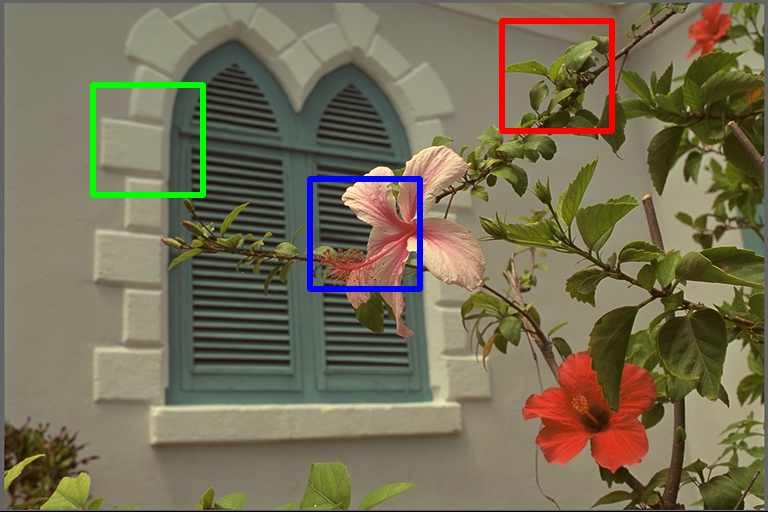} &
 \includegraphics[width=.08695656427217936\linewidth]{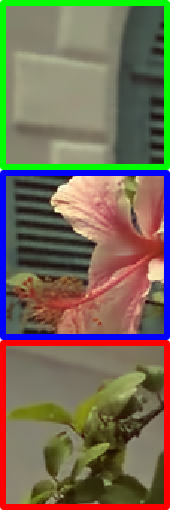} & 
 \includegraphics[width=.08695656427217936\linewidth]{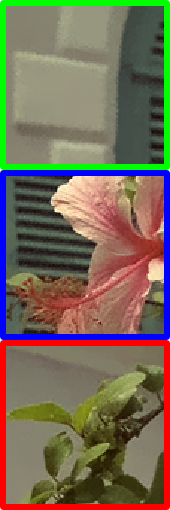} & 
 \includegraphics[width=.08695656427217936\linewidth]{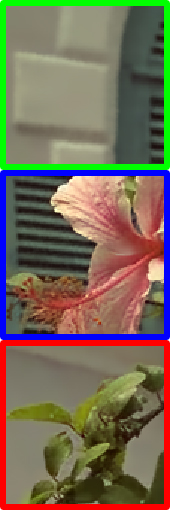} & 
 \includegraphics[width=.08695656427217936\linewidth]{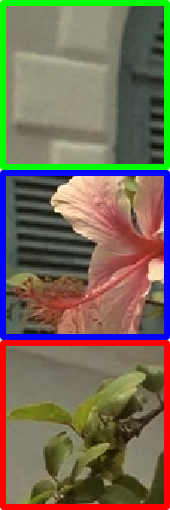} &
 \includegraphics[width=.08695656427217936\linewidth]{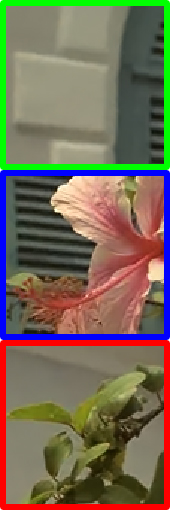} &
 \includegraphics[width=.08695656427217936\linewidth]{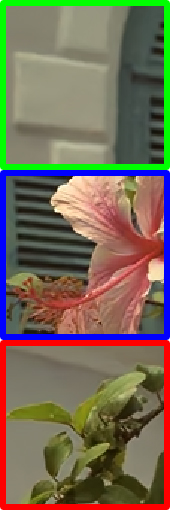} & 
 \includegraphics[width=.08695656427217936\linewidth]{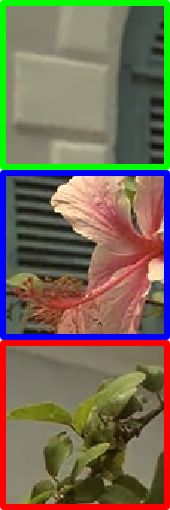} \\
  \multirow{2}{*}{\footnotesize{Target image}} & \footnotesize{$\mc S_p^p$} & \footnotesize{$\mc S_1^{\bm w}$} & \footnotesize{$\mc S_p^{\bm w}$} & \footnotesize{$\ell_1$} & \footnotesize{$\ell_p^p$} & \footnotesize{$\ell_1^{\bm w}$} & \footnotesize{$\ell_p^{\bm w}$}\\
 & \footnotesize{34.10} & \footnotesize{33.55} & \footnotesize{34.92} & \footnotesize{35.13} & \footnotesize{35.70} & \footnotesize{\textbf{37.02}} & \footnotesize{36.11} \\
\end{tabular}
   \caption{Visual comparisons among several methods on a mosaicked sample from Kodak dataset with 3\% noise (best viewed with x5 magnification). For each demosaicked image its PSNR value is provided in dB.}
   \label{fig:DemosaicSynthetic3pn}
\end{figure*}

\begin{figure*}[t]
\centering
\setlength\extrarowheight{-5pt}
\begin{tabular}{@{} c @{} c @{} c @{} c @{} c @{} c @{} c @{} c @{}}
  \multirow{3}{*}[1.85cm]{\includegraphics[width=.244\linewidth]{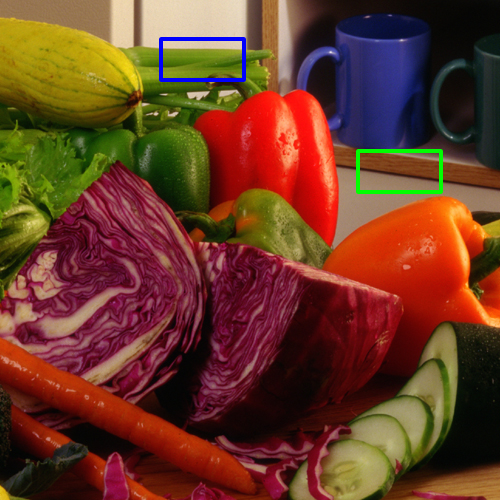}} &
 \includegraphics[width=.108\linewidth]{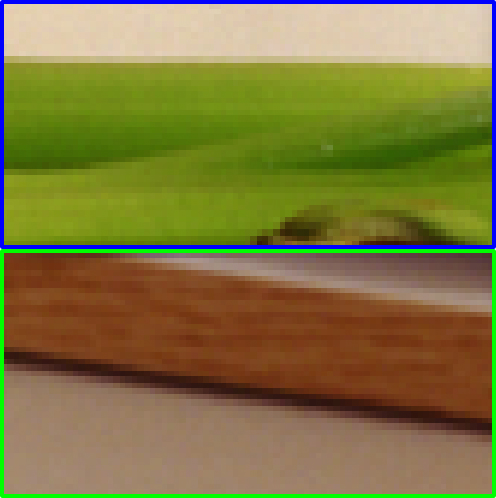} & 
 \includegraphics[width=.108\linewidth]{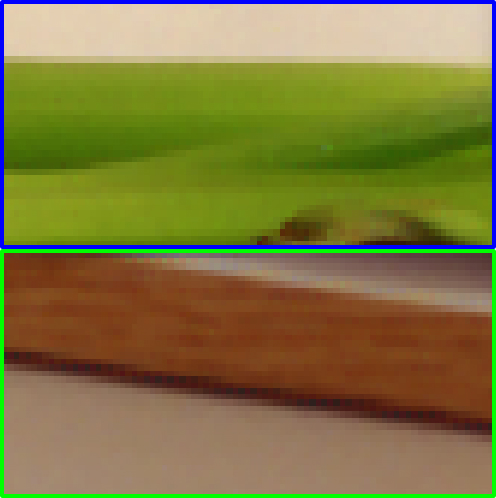} & 
 \includegraphics[width=.108\linewidth]{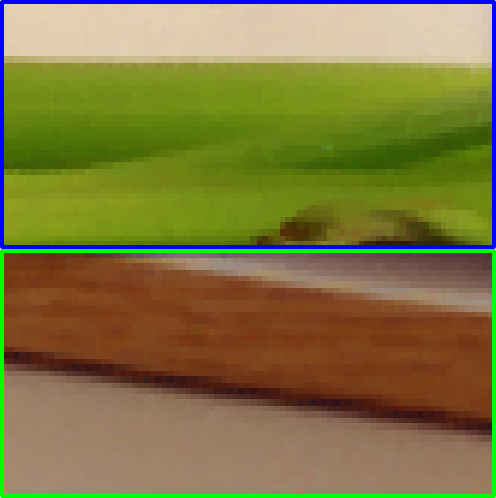} & 
 \includegraphics[width=.108\linewidth]{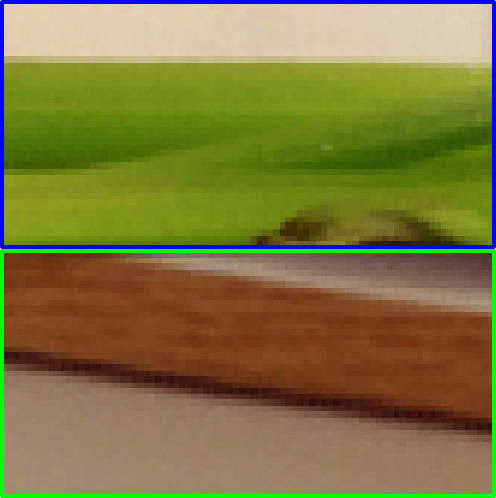} &
 \includegraphics[width=.108\linewidth]{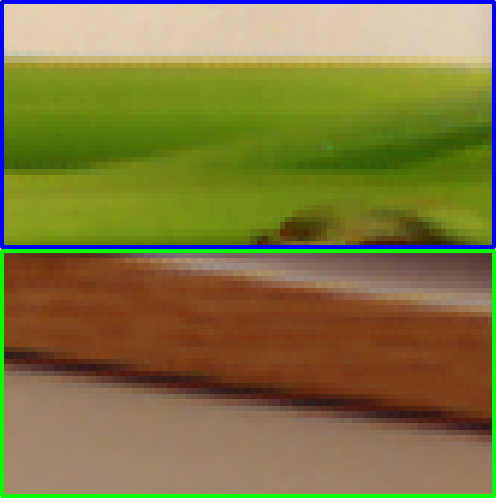} &
 \includegraphics[width=.108\linewidth]{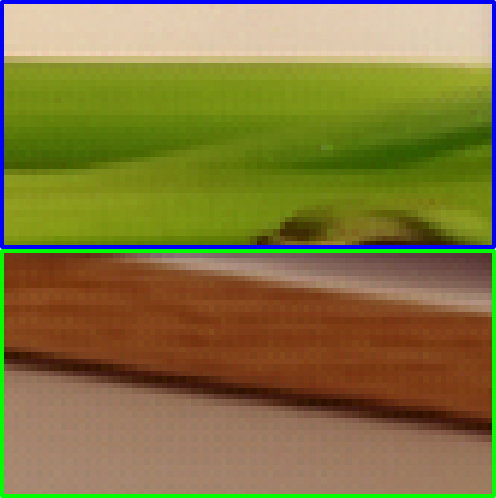} & 
 \includegraphics[width=.108\linewidth]{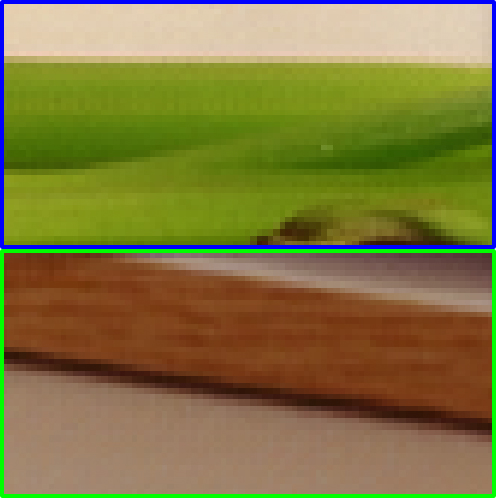} \\
 
 & \footnotesize{Target} & \footnotesize{VTV: 35.16} & \footnotesize{TVN: 36.21} & \footnotesize{$\mc S_1$: 38.18} & \footnotesize{Bilinear: 35.24} & \footnotesize{RED: 36.42} & \footnotesize{IRCNN: 39.57}\\
 & \includegraphics[width=.108\linewidth]{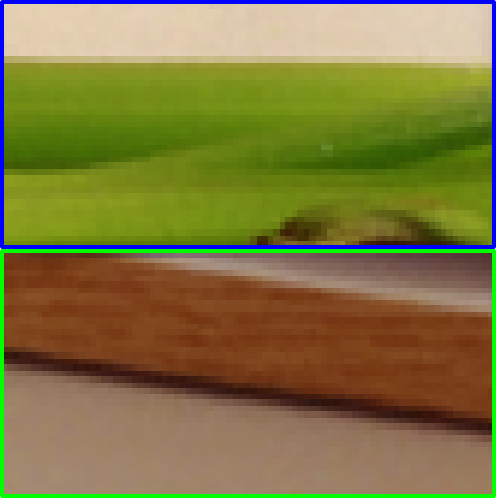} & 
 \includegraphics[width=.108\linewidth]{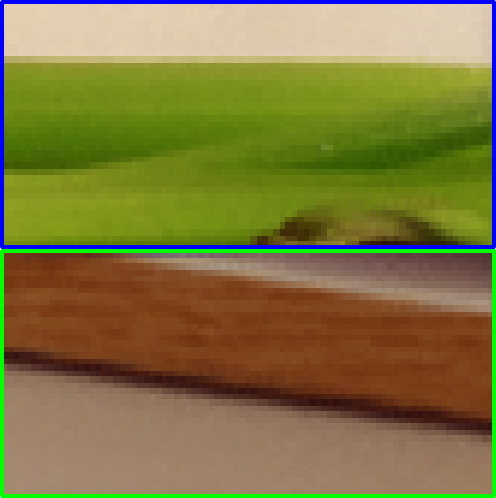} & 
 \includegraphics[width=.108\linewidth]{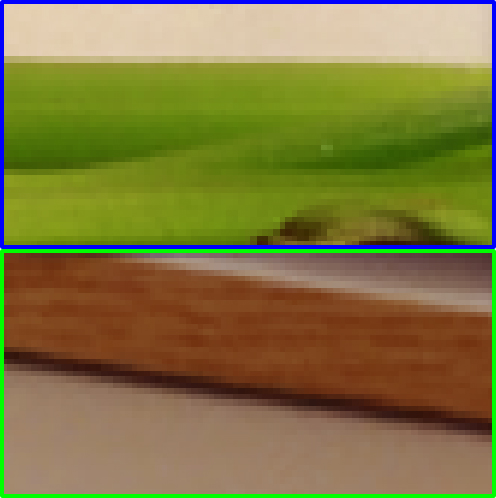} & 
 \includegraphics[width=.108\linewidth]{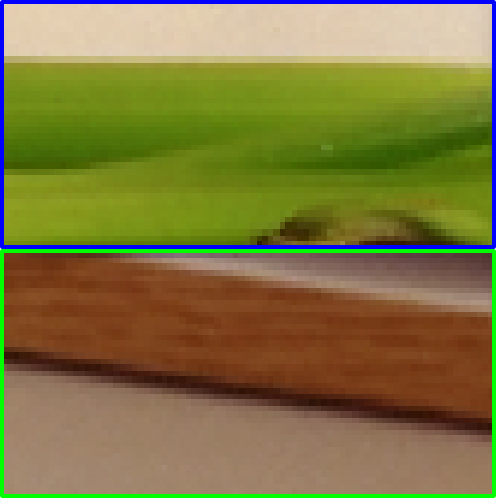} &
 \includegraphics[width=.108\linewidth]{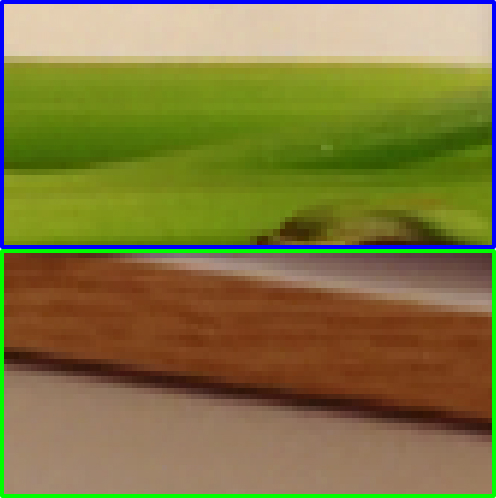} &
 \includegraphics[width=.108\linewidth]{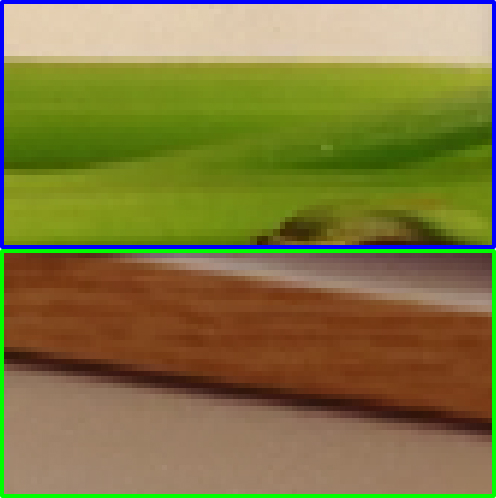} & 
 \includegraphics[width=.108\linewidth]{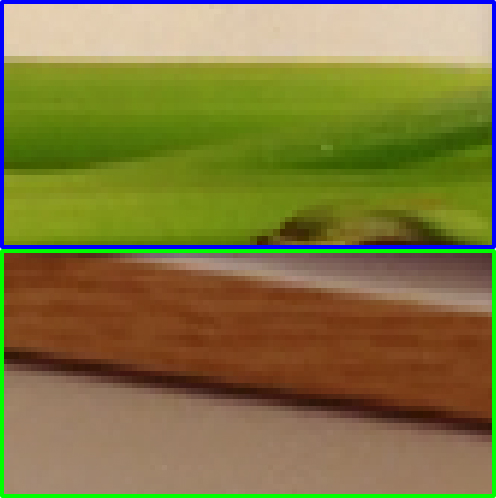}\\
  \footnotesize{Target image} & \footnotesize{$\mc S_p^p$: \textbf{39.97}} & \footnotesize{$\mc S_1^{\bm w}$: 38.35} & \footnotesize{$\mc S_p^{\bm w}$: 39.83} & \footnotesize{$\ell_1$: 37.91} & \footnotesize{$\ell_p^p$: 38.84} & \footnotesize{$\ell_1^{\bm w}$: 39.40} & \footnotesize{$\ell_p^{\bm w}$: 38.79}\\
\end{tabular}
   \caption{Visual comparisons among competing methods on a sample from McMaster dataset~\cite{Zhang2011} (best viewed with x5 magnification) in a demosaick only scenario. For each reconstructed image its PSNR value is provided in dB.}
\label{fig:DemosaicSynthetic0pn}
\end{figure*}

\textbf{Demosaicking.} For this recovery task we compare our low-rank and sparsity promoting LIRLS models against the same general-purpose classical and deep learning approaches as in other two problems. In addition we consider the bilinear interpolation method. The average results of all the competing methods are reported in Table~\ref{tab:demosaic_color}, and the visual comparisons among the different methods are provided in Fig.~\ref{fig:DemosaicSynthetic3pn} for the high noise case and in Fig.~\ref{fig:DemosaicSynthetic0pn} for the low one. From these results we also come to similar conclusions as with the previous problems. Specifically, the best performance on average is achieved by the $\ell_1^{\bm w}$ LIRLS model. The only different behavior that we observe in this problem is is that on the McMaster dataset the $\mc S_p^{p}$ and $\mc S_p^{p, \bm w}$ models in general outperform the $\ell_1^{\bm w}$ one.

We have found that this effect can be attributed to the significant differences in the generation process of the images in the datasets. Indeed, as was mentioned in previous works~\cite{Li2008,Zhang2011}, the Kodak dataset consists of scanned film-based photos, which have low saturation and are smooth in the chromatic gradient. Compared to them, the McMaster data consist of IMAX images, which are more saturated and better resemble images acquired by  digital cameras. We should recall, that the images we used for training are taken from the BSD500 image dataset consisting of the Corel Stock Photos which originate from film-based photos, so it should be expected our models to show a better performance on the Kodak dataset. Furthermore, this effect should be less pronounced for the images altered by heavy noise, which equalizes the distributions of two different image sets. All these notions are fully confirmed by our experimental results. Thus, the best performing models on McMaster dataset with lower noise levels are those that do not contain the weight prediction network with the best performing being the $\mc S_p^p$ penalty, while with the increase of the noise levels up to 3\% the $\mc S_1^{\bm w}, \mc S_p^{\bm w}, \ell_1^{\bm w}, \ell_p^{\bm w}$ models gradually start outperforming it. At the same time on the Kodak dataset all these models outperform the $\mc S_p^p$ one across all noise levels.

\section{Conclusions}
In this work we have extended our recently proposed learned IRLS method, which covers a rich family of sparsity and low-rank promoting priors. We have provided a formal proof of its linear convergence to a stationary point and derived an upper bound for the convergence rate. Our extensive experiments targeted color image restoration problems that further extended our previously published results. These results show that our method can be successfully applied to deal with a wide range of practical inverse problems under a wide range of degradations and noise levels. In addition, we have managed to learn our sparse and low-rank regularizers from data in the context of bilevel optimization, which thanks to the provided convergence guarantees of our minimization approach boils down to supervised learning with implicit backpropagation. As a result, despite the fact that all our models involve only a small number of parameters, they still lead to very competitive reconstructions. Given the very good performance of our learned priors, an interesting open research topic is to further accelerate the inference procedure by either utilizing other optimization techniques or by constructing dedicated preconditioners for each specific reconstruction problem. Another possible research direction would be to apply the proposed techniques to other restoration problems, e.g. inpainting, colorization, etc. Finally, an interesting question that could be further investigated would be whether the proposed learned priors could act as a good parametric form for modelling distributions of natural images.

\appendix[Proof of Proposition~\ref{prop:convergence}]\label{sec:prop_proof}
It is straightforward to show by direct differentiation that the gradients of the objective functions and their majorizers, in both sparse and low-rank cases, could be written as:
\bal
\nabla \mc J\pr{\bm x^k} =& \tfrac{1}{\sigma_{\bm n}^2}\bm A^\transp \pr{\bm A \bm x^k - \bm y} + p \suml_{i=1}^\ell \bm G_i^\transp \bm W_i^k  \bm G_i \bm x^k = \nonumber \\ &\tfrac{1}{\sigma_{\bm n}^2} \pr{\bm S^k \bm x^k - \bm A^\transp \bm y}
\label{eq:grad_majorizer}
\eal
\bal
\nabla \tilde{\mc Q}\pr{\bm x; \bm x^k} =& \tfrac{1}{\sigma_{\bm n}^2}\bm A^\transp \pr{\bm A \bm x - \bm y} + \nonumber \\ &p \suml_{i=1}^\ell \bm G_i^\transp \bm W_i^k  \bm G_i \bm x + \delta \pr{\bm x - \bm x^k},
\eal
where $\bm W_i^k = \begin{cases}\bm W_{\bm z_i^k} \\ \bm I_q \otimes \bm W_{\bm Z_i^k} \end{cases}$\!\!\!\!\!. From this result we can see, that the condition $\nabla \tilde{\mc Q} \pr{\bm x^k;\bm x^k} = \nabla \mc J\pr{\bm x^k}$ from \eqref{eq:additional_MM_properties} is satisfied. The second condition is also satisfied, since we have specifically crafted our majorizers to be quadratic, hence continuous. As a result, the MM framework guarantees convergence of our algorithm to a stationary point for both sparse and low-rank functionals, and we will show this further using other way of reasoning. Another fact, that is easy to prove by direct substitution and which will be found useful is that: 
\bal
\label{eq:majorizer_taylor_like}
\tilde{\mc Q} \pr{\bm x; \bm x^k} = \mc J\pr{\bm x^k} + \pr{\bm x - \bm x^k}^\transp \nabla \mc J\pr{\bm x^k} + \nonumber \\ \tfrac{1}{2\sigma_{\bm n}^2} \norm{\bm x - \bm x^k}{\bm S^k + \alpha \bm I}^2.
\eal
The proofs provided below are related to the convergence of our algorithm and are similar in spirit with those provided by~\cite{Chaudhury2013} for the task of non-local patch-based denoising. 

\subsection{Convergence to Stationary Point}
At first, we consider two successive iterations $\bm x^k$ and $\bm x^{k + 1}$. By construction $\bm x^{k + 1}$ delivers the minimum of $\tilde{\mc Q}\pr{\bm x; \bm x^k}$, meaning $\nabla \tilde{\mc Q}\pr{\bm x^{k + 1}; \bm x^k} = 0$. Using Eq.~\ref{eq:majorizer_taylor_like} we obtain:
\bal
\nabla \tilde{\mc Q}\pr{\bm x^{k + 1}; \bm x^k} =& \nabla \mc J\pr{\bm x^k} + \nonumber\\ &\tfrac{1}{\sigma_{\bm n}^2} \pr{\bm S^k + \alpha \bm I}\pr{\bm x^{k+1} - \bm x^k} = 0,
\eal
which implies that it holds:
\bal
\label{eq:grad_objective_short}
\nabla \mc J\pr{\bm x^k} = - \tfrac{1}{\sigma_{\bm n}^2} \pr{\bm S^k + \alpha \bm I}\pr{\bm x^{k+1} - \bm x^k}.
\eal
We use this together with the majorizing property to write:
\bal
\mc J\pr{\bm x^{k+1}} &\le \tilde{\mc Q}\pr{\bm x^{k+1}; \bm x^k} = \mc J\pr{\bm x^k} + \nonumber\\&\pr{\bm x^{k+1} - \bm x^k}^\transp \nabla \mc J\pr{\bm x^k} + \tfrac{1}{2\sigma_{\bm n}^2} \norm{\bm x^{k+1} - \bm x^k}{\bm S^k + \alpha \bm I}^2 \nonumber \\ &= \mc J\pr{\bm x^k}  - \tfrac{1}{2\sigma_{\bm n}^2} \norm{\bm x^{k+1} - \bm x^k}{\bm S^k + \alpha \bm I}^2.
\eal
From the above it directly follows that: 
\bal
\label{eq:x_j_ineq_start}
\norm{\bm x^{k+1} - \bm x^k}{\bm S^k + \alpha \bm I}^2 \le 2\sigma_{\bm n}^2 \pr{\mc J\pr{\bm x^k} - \mc J\pr{\bm x^{k+1}}}.
\eal
The left hand side of the inequality above can be represented as:
\bal
\label{eq:norms_decomp}
&\norm{\bm x^{k+1} - \bm x^k}{\bm S^k + \alpha \bm I}^2 = \norm{\bm x^{k+1} - \bm x^k}{\bm A^\transp \bm A}^2 + \nonumber\\&\quad p \sigma_{\bm n}^2 \suml_{i=1}^\ell \norm{\bm x^{k+1} - \bm x^k}{\bm G_i^\transp \bm W_i^k \bm G_i}^2 + \alpha \norm{\bm x^{k+1} - \bm x^k}{2}^2.
\eal
We now point out the fact, that for any (possibly complex) matrix $\bm B \in \C^{m\times n}$, there exists an eigendecomposition of the form  $\bm B^H \bm B = \bm U_{\bm B} \bm \Lambda_{\bm B} \bm U^H_{\bm B}$, where $\bm U_{\bm B}\in \C^{n\times n}$ is the unitary matrix of eigenvectors, and $\Lambda_{\bm B} = \diag{\bm \lambda_{\bm B}}$ with the vector of eigenvalues $\bm \lambda_{\bm B}\in \R^{n}_+$. This allows us for any vector $\bm v \in \C^{n}$ to write $\norm{\bm v}{\bm B^H \bm B}^2 = \norm{\bm U \bm v}{\bm \Lambda_{\bm B}}^2 \ge \lambda_{\min}\pr{\bm B^H \bm B} \norm{\bm U_{\bm B} \bm v}{2}^2 =  \lambda_{\min}\pr{\bm B^H \bm B} \norm{\bm v}{2}^2$, where we have used the norm-preserving property of the unitary matrix $\bm U_{\bm B}$. Applying this on Eq.~\eqref{eq:norms_decomp} we get: 
\bal
&\norm{\bm x^{k+1} - \bm x^k}{\bm S^k + \alpha \bm I}^2 \ge \nonumber\\&\underbrace{\br{\lambda_{\min}\pr{\bm A^H \bm A} + p \sigma_{\bm n}^2 \gamma^{\frac{p-2}{2}} \suml_{i=1}^\ell \begin{cases}\bm w_i^{\min} \\ \bm w_{i1}\end{cases}\!\!\!\!\! \lambda_{\min}\pr{\bm G_i^\transp \bm G_i} + \alpha}}_{\beta} \nonumber\\&\quad\quad\cdot \norm{\bm x^{k+1} - \bm x^k}{2}^2. \label{eq:xdiff_mahalanobis_to_norm}
\eal 
For the low-rank case the weights $\bm w$ are sorted in increasing order, so it always holds $\bm w_i^{\min} = \bm w_{i1}$. In the case of ill-posed reconstruction problems (which we are dealing with in this work) it always holds $\lambda_{\min}\pr{\bm A^H \bm A} = 0$. In our derivations we preserve this term in order to cover the non-singular cases. Substituting Eq.~\eqref{eq:xdiff_mahalanobis_to_norm} into Eq.~\eqref{eq:x_j_ineq_start} we get: 
\bal
\norm{\bm x^{k+1} - \bm x^k}{2}^2 \le \frac{2\sigma_{\bm n}^2}{\beta} \pr{\mc J\pr{\bm x^k} - \mc J\pr{\bm x^{k+1}}}.
\eal
Now, it is straightforward to show the convergence of the sequence $\bm x^k$:
\bal
0 \le \lim_{t \to \infty} \norm{\bm x^{k+1} - \bm x^k}{2}^2 \le \lim_{t \to \infty} \frac{2\sigma_{\bm n}^2}{\beta} \pr{\mc J\pr{\bm x^k} - \mc J\pr{\bm x^{k+1}}} = 0,
\eal
which according to the squeeze theorem it implies that $\lim_{t \to \infty} \norm{\bm x^{k+1} - \bm x^k}{2}^2 = 0$. It is now easy to show that $\bm x^* = \lim_{t \to \infty} \bm x^k$ is also a stationary point of $\mc J\pr{\bm x}$, \ie $\cbr{\bm x^k}$ converges to a stationary point of $\mc J\pr{\bm x}$. Taking the limit of Eq.~\eqref{eq:grad_objective_short}, we get:
\bal 
&\nabla \mc J\pr{\bm x^*} = -\tfrac{1}{\sigma_{\bm n}^2}\bm A^\transp \bm A \pr{\bm x^* - \bm x^*}\nonumber\\&\quad -p \suml_{i=1}^\ell \bm G_i^\transp \bm W_i^*  \bm G_i \pr{\bm x^* - \bm x^*} - \delta \pr{\bm x^* - \bm x^*} = 0,
\eal
which suggests that $\bm x^*$ is a stationary point of $\nabla \mc J\pr{\bm x}$. This completes the proof of the first item in Proposition \ref{prop:convergence}.

\subsection{Convergence rate}
To compute the convergence rate of the value of the objective function $\mc J\pr{\bm x}$, we proceed with the comparison of this function at the points $\bm x^{k+1}$ and $\theta_k \bm x^k + \pr{1 - \theta_k} \bm x^*$ with $\theta_k$ %$0 < \theta_k < 1$ 
being some constant. Using the majorization property of $\tilde{\mc Q}$ and the optimality property of $\bm x^{k+1}$ in $\tilde{\mc Q}\pr{\bm x^{k+1}; \bm x^k}$ we can write:
\bal
\mc J\pr{x^{k+1}} \le \tilde{\mc Q}\pr{\bm x^{k+1}; \bm x^k} \le \tilde{\mc Q}\pr{\theta_k \bm x^k + \pr{1 - \theta_k} \bm x^*; \bm x^k}.
\eal
We now use Eq.~\eqref{eq:majorizer_taylor_like} to get:
\bal
&\tilde{\mc Q}\pr{\theta_k \bm x^k + \pr{1 - \theta_k} \bm x^*; \bm x^k} = \mc J\pr{\bm x^k} + \nonumber\\ &\pr{1 - \theta_k}\pr{\bm x^* - \bm x^k}^\transp \nabla \mc J\pr{\bm x^k} + \frac{\pr{1-\theta_k}^2}{2\sigma_{\bm n}^2} \norm{\bm x^* - \bm x^k}{\bm S^k + \alpha \bm I}^2.
\eal
At the same time it holds:
\bal
\tilde{\mc Q}\pr{\bm x^*; \bm x^k} = \mc J\pr{\bm x^k} + \pr{\bm x^* - \bm x^k}^\transp \nabla \mc J\pr{\bm x^k} + \nonumber\\\tfrac{1}{2\sigma_{\bm n}^2} \norm{\bm x^* - \bm x^k}{\bm S^k + \alpha \bm I}^2.
\label{eq:maj_in_stationary_point}
\eal
Combining the two, we get:
\bal
&\tilde{\mc Q}\pr{\theta_k \bm x^k + \pr{1 - \theta_k} \bm x^*; \bm x^k} = \theta_k \mc J\pr{\bm x^k} + \nonumber\\&\quad\pr{1 - \theta_k}\br{\tilde{\mc Q}\pr{\bm x^*; \bm x^k} - \frac{\theta_k}{2\sigma_{\bm n}^2}\norm{\bm x^* - \bm x^k}{\bm S^k + \alpha \bm I}^2}.
\eal
Let us now consider $0 \le \theta_k = \frac{2\sigma_{\bm n}^2 \pr{\tilde{\mc Q}\pr{\bm x^*; \bm x^k} - \mc J\pr{\bm x^*}}}{\norm{\bm x^* - \bm x^k}{\bm S^k + \alpha \bm I}^2}$. Under this assumption we have:
\bal
\mc J\pr{\bm x^{k+1}} \le& \tilde{\mc Q}\pr{\bm x^{k+1}; \bm x^k} \le \tilde{\mc Q}\pr{\theta_k \bm x^k + \pr{1 - \theta_k} \bm x^*; \bm x^k} \nonumber\\=&\theta_k \mc J\pr{\bm x^k} + \mc J\pr{\bm x^*} - \theta_k \mc J\pr{\bm x^*}, 
\eal
which translates to:
\bal
\mc J\pr{\bm x^{k+1}} - \mc J\pr{\bm x^*} \le \theta_k \pr{\mc J\pr{\bm x^k} - \mc J\pr{\bm x^*}}.
\eal

We then use the Taylor's theorem with the Lagrange remainder form to expand $\mc J\pr{\bm x}$ at the point $\bm x^*$ near the point $\bm x^k$:
\bal
\mc J\pr{\bm x^*} =& \mc J\pr{\bm x^k} + \pr{\bm x^* - \bm x^k}^\transp \nabla \mc J\pr{\bm x^k}\nonumber\\&+\frac{1}{2} \pr{\bm x^* - \bm x^k}^\transp \bm H_{\mc J\pr{\tilde{\bm x}^k}} \pr{\bm x^* - \bm x^k},
\eal
where $\tilde{\bm x}^k$ lies in the segment between $\bm x^*$ and $\bm x^k$. Such expansion together with Eq.~\eqref{eq:maj_in_stationary_point} allows us to write:
\bal
\theta_k &= 1 - \sigma_{\bm n}^2 \frac{\norm{\bm x^* - \bm x^k}{\bm H_{\mc J\pr{\tilde{\bm x}^k}}}^2}{\norm{\bm x^* - \bm x^k}{\bm S^k + \alpha \bm I}^2} \nonumber\\&\le 1 - \sigma_{\bm n}^2 \lambda_{\min}\pr{ \bm H_{\mc J\pr{\tilde{\bm x}^k}}} \frac{\norm{\bm x^* - \bm x^k}{2}^2}{\norm{\bm x^* - \bm x^k}{\bm S^k + \alpha \bm I}^2}.
\eal
Next, we write: 
\bal
&\norm{\bm x^* - \bm x^k}{\bm S^k + \alpha \bm I}^2 = %\norm{\bm x^* - \bm x^k}{\bm A^\transp \bm A}^2 + p \sigma_{\bm n}^2 \suml_{i=1}^\ell \norm{\bm x^* - \bm x^k}{\bm G_i^\transp \bm W_i^k \bm G_i}^2 + \alpha \norm{\bm x^* - \bm x^k}{2}^2 = \\ = 
\norm{\bm A\pr{\bm x^* - \bm x^k}}{2}^2 + \nonumber\\&\quad\quad p\sigma_{\bm n}^2 \suml_{i=1}^\ell \norm{{\bm W_i^k}^{1/2} \bm G_i\pr{\bm x^* - \bm x^k}}{2}^2 + \alpha \norm{\bm x^* - \bm x^k}{2}^2 \nonumber\\ %\le \br{\norm{\bm A}{2}^2 + p \sigma_{\bm n}^2 \suml_{i=1}^\ell \norm{{\bm W_i^k}^{1/2} \bm G_i}{2}^2 + \alpha} \norm{\bm x^* - \bm x^k}{2}^2 \le \\ 
&\le \br{\norm{\bm A}{2}^2 + p \sigma_{\bm n}^2 \suml_{i=1}^\ell \norm{{\bm W_i^k}^{1/2}}{2}^2 \norm{\bm G_i}{2}^2 + \alpha} \norm{\bm x^* - \bm x^k}{2}^2.
\eal
Using this result, we can further transform the upper bound for $\theta_k$ as:
\bal
0 \le \theta_k \le 1 - \frac{\sigma_{\bm n}^2 \lambda_{\min}\pr{ \bm H_{\mc J\pr{\tilde{\bm x}^k}}}}{\norm{\bm A}{2}^2 + p \sigma_{\bm n}^2 \suml_{i=1}^\ell \norm{{\bm W_i^k}^{1/2}}{2}^2 \norm{\bm G_i}{2}^2 + \alpha}.
\eal
From this we can derive the convergence regime for $\mc J\pr{\bm x}$ and an upper bound for the convergence rate $\nu_{\mc J}$:
\bal
\nu_{\mc J} \equiv &\lim_{t \to \infty}{\frac{\mc J\pr{\bm x^{k+1}} - \mc J\pr{\bm x^*}}{\mc J\pr{\bm x^k} - \mc J\pr{\bm x^*}}} \le \lim_{t \to \infty}{\theta_k} \le \nonumber\\%1 - \lim_{t \to \infty}{\frac{\sigma_{\bm n}^2 \lambda_{\min}\pr{ \bm H_{\mc J\pr{\tilde{\bm x}^k}}}}{\norm{\bm A}{2}^2 + p \sigma_{\bm n}^2 \suml_{i=1}^\ell \norm{{\bm W_i^k}^{1/2}}{2}^2 \norm{\bm G_i}{2}^2 + \alpha}} = \\ = 
&1 - \frac{\sigma_{\bm n}^2 \lambda_{\min}\pr{ \bm H_{\mc J\pr{\bm x^*}}}}{\norm{\bm A}{2}^2 + p \sigma_{\bm n}^2 \suml_{i=1}^\ell \norm{{\bm W_i^*}^{1/2}}{2}^2 \norm{\bm G_i}{2}^2 + \alpha}.
\eal
We note that we can calculate the norm of $\bm W_i^*$ as follows:
\bal
\norm{{\bm W_i^*}^{1/2}}{2}^2 = \lambda_{\max} \pr{\bm W_i^*} = \begin{cases} \max_j{\bm w_{ij} \pr{{\bm z_i^*}_j^2 + \gamma}^{\frac{p-2}{2}}} \\ \max_j{\bm w_{ij} \pr{\bm \sigma_j^2\pr{\bm Z_i^*} + \gamma}^{\frac{p-2}{2}}}\end{cases},
\eal 
which allows us to finally write our upper bound for the convergence rate as:
\bal
&\nu_{\mc J}\le \nu_{\mc J}^{ub} \equiv 1 - \nonumber\\&\frac{\sigma_{\bm n}^2 \lambda_{\min}\pr{ \bm H_{\mc J\pr{\bm x^*}}}}{\norm{\bm A}{2}^2 + p \sigma_{\bm n}^2 \suml_{i=1}^\ell \norm{\bm G_i}{2}^2 \begin{cases} \max_j{\bm w_{ij} \pr{{\bm z_i^*}_j^2 + \gamma}^{\frac{p-2}{2}}} \\ \max_j{\bm w_{ij} \pr{\bm \sigma_j^2\pr{\bm Z_i^*} + \gamma}^{\frac{p-2}{2}}}\end{cases}\!\!\!\!\!\! + \alpha}.
\eal
Since the MM framework ensures the monotonicity of $\cbr{\mc J\pr{\bm x^k}}$, $\bm x^\ast$ can not be a local maximum, so $\bm H_{\mc J\pr{\bm x^*}}$ is either positive-definite or indefinite. The result we have obtained implies that once our iterative process enters the region sufficiently close to the local minima of $\mc J\pr{\bm x}$, then $\lambda_{\min}\pr{ \bm H_{\mc J\pr{\bm x^*}}} > 0$ meaning $\nu_{\mc J} < 1$. This leads to the fact that from this point the convergence towards this local minima is linear with a convergence rate of at least $\nu_{\mc J}^{ub}$. 

\bibliographystyle{IEEEtran}
\bibliography{main}
\end{document}